\documentclass{article}
\usepackage[utf8]{inputenc}
\usepackage{amsthm}
\usepackage{amsmath,amssymb}
\usepackage[numbers,sort&compress]{natbib}
\usepackage[T1]{fontenc}    
\usepackage{hyperref}       
\usepackage{url}            
\usepackage{booktabs}       
\usepackage{amsfonts}       
\usepackage{nicefrac}       
\usepackage{microtype}      
\usepackage{graphicx}
\usepackage{placeins}

\usepackage{authblk}

\usepackage{xcolor, soul}
\usepackage{color, colortbl}
\definecolor{Gray}{gray}{0.8}
\definecolor{LightCyan}{rgb}{0.88,1,1}
\definecolor{green(html/cssgreen)}{rgb}{0.0, 0.5, 0.0}

\hypersetup{
    colorlinks=true,
    linkcolor=black,
    citecolor=black,      
    urlcolor=blue
}

\usepackage[font=small,format=plain,labelfont=bf]{caption} 
\usepackage{array} 
\usepackage[rightcaption]{sidecap}

\newtheorem{remark}{Remark}
\newtheorem{remark*}{Remark}
 
\newtheorem{theorem*}{Theorem}

\usepackage{mathtools}
\usepackage{mdframed}

\usepackage[CaptionBefore]{fltpage}
\usepackage{comment}

\usepackage[margin=1in]{geometry}
\usepackage{enumitem}

\newmdtheoremenv{theo}{Theorem}

\usepackage{accents}
\usepackage{bbm}

\let\oldstar\*
\renewcommand{\*}[1]{\mathbf{#1}}

\newcommand{\x}{\mathbf{x}}

\newcommand{\W}{\mathbf{W}}

\newcommand{\fR}{\mathbb{R}}

\newcommand*{\dt}[1]{\accentset{\raisebox{1pt}{\scalebox{0.4}{$\bullet$}}}{#1}}

\usepackage[most]{tcolorbox}

\title{Contraction Properties of the Global Workspace Primitive}
\author[1,3,*]{\textbf{Michaela Ennis}}
\author[1,*]{\textbf{Leo Kozachkov}}
\author[1,2]{\textbf{Jean-Jacques Slotine}}

\affil[1]{Department of Brain and Cognitive Sciences, Massachusetts Institute of Technology}
\affil[2]{Department of Mechanical Engineering, Massachusetts Institute of Technology}
\affil[3]{Division of Medical Sciences, Harvard University}
\affil[*]{\textbf{Equal contribution}}
\affil[ ]{\texttt{\{mennis,leokoz8,jjs\}@mit.edu}}

\date{October 2023}

\begin{document}

\renewcommand{\equationautorefname}{}

\maketitle

\begin{abstract}
\noindent To push forward the important emerging research field surrounding multi-area recurrent neural networks (RNNs), we expand theoretically and empirically on the provably stable "RNNs of RNNs" introduced by \citet{kozachkov2022rnns}. We prove relaxed stability conditions for salient special cases of this architecture, most notably for a global workspace modular structure. We then demonstrate empirical success for "Global Workspace Sparse Combo Nets" with a small number of trainable parameters, not only through strong overall test performance but also greater resilience to removal of individual subnetworks. These empirical results for the global workspace \emph{inter}-area topology are contingent on stability preservation, highlighting the relevance of our theoretical work for enabling modular RNN success. Further, by exploring sparsity in the connectivity structure between different subnetwork modules more broadly, we improve the state of the art performance for stable RNNs on benchmark sequence processing tasks, thus underscoring the general utility of specialized graph structures for multi-area RNNs. 
\end{abstract}

\vspace{0.75cm}

\noindent Most of the recent paradigm shifts in deep learning have resulted from novel architectural designs, designs often motivated by similarities to hypotheses about the brain \citep{sejnowski}. Imposing hard constraints on feedforward network structure lead to a number of these major advances, including the convolutional neural network (CNN). Like CNNs, RNNs have a history in computational neuroscience well preceding their widespread application to practical machine learning problems. Despite this, and the obvious presence of \textit{modular} recurrent structures in the brain, there has not been much parallel development of RNN architectures; even within neuroscience theory, multi-area networks remain a niche field of study \citep{yang2021towards}. 

In contrast with the success of Transformers for processing large scale sequence data \citep{transformers}, a major challenge for the scalability of recurrent neural networks has been \textit{stability} during learning \citep{miller2018stable}. Fortunately, modularity is an extremely useful principle for building up large stable systems. By combining control theory tools with salient multi-area RNN topologies, we can generate a vast space of provably stable RNN model designs that possess a number of desirable properties for engineered systems. 

Ultimately, there is increasing demand for deep learning agents that can accurately assess uncertainty, behave in an interpretable way, continually learn, and so forth. It is highly plausible then that some flavor of multi-area RNN architecture, enabled by a stable connectivity structure, will eventually cause a strong shift in practice for one or more areas of machine learning that remain especially challenging today. To further this aim, we highlight the variety of benefits that can come with the modularity and stability of "RNNs of RNNs" like described by \citet{kozachkov2022rnns}, and contribute additional characterization of as well as improvements to this class of model architectures, both theoretically and empirically. Taking inspiration from neuroscience, we focus on a stable multi-area RNN structure derived from the global workspace theory of cognition \citep{newell1972human,baars1993cognitive,dehaene1998neuronal}.

\section{Neurobiological Motivations}
\label{sec:neuro-intro}
Modularity is a core design principle across many fields of engineering and computer science. While abstraction of details may be messier when describing biological systems, they are highly modular by nature, on scales from molecular to cellular to organismal biology \citep{Lorenz2011}. Moreover, the existence of specialized roles for different regions of the brain has been well established within neuroscience for a long time \citep{kandel2000principles}, and phenomena thought to be important to neural computation like traveling waves have been observed on both the whole brain and single area level \citep{Muller2018}. It is thus quite surprising that in the highly interdisciplinary domain of artificial neural networks, very little research on modular architectures exists. 

While technical and computational limitations drove early experimental neurobiology research to the study of individual brain areas in isolation, 
empirical work involving multi-area recordings has become increasingly widespread \citep{mashour2020conscious,Abbott_Svoboda_2020,Perich_2021,semedo2019cortical,yang2021towards}. Indeed, \citet{michaels2020goal} recently showed that modular neural networks were much more capable of predicting parietofrontal cortex activity during a macaque grasping task than traditional neural networks were, suggesting a multi-area approach will be important for accurately analyzing complex neural recording data. RNN theory needs to adapt accordingly as open questions increase in complexity at a staggering rate. In particular, much remains unknown about how different regions of the brain coordinate to accomplish broader goals. Gaining a better understanding of the connections \emph{between} subnetworks in a multi-area RNN framework is therefore a primary goal of our work. Because neurons in the cortex receive the majority of their inputs from local \emph{intra-}area synapses \citep{Muller2018}, we are especially interested in investigation of sparse structures for \emph{inter-}area weights.

From an evolutionary biology perspective, a greater research focus on connections between modules would also be well-justified. There is strong evidence for the facilitated variation mechanism of evolution, which proposes that the majority of traits developed in the last $\sim 400$ million years were the result of evolutionary changes to regulatory elements that combine "core modules" rather than changes to those modules themselves \citep{gerhart2007fv,parter2008fv}. More generally, modularity confers a number of evolutionary benefits. In a modular system, one component can be changed without perturbation of other components, and duplication (plus subsequent variation of) existing modules can allow much more efficient expansion to new functionalities \citep{Lorenz2011}. This is especially true in the natural world, where the environment may change in a fashion that shifts goals to be composed of different combinations of subtasks at different times. Computational evolutionary algorithms will in fact lead to modular model structures when implemented in a regime with modular task switching \citep{Kashtan2005}. For these reasons, the \emph{inter-}area connectivity patterns that may emerge under different learning algorithms and for different tasks are another area of major interest.

\subsection{Global Workspace Theory}
\label{subsubsec:gw-intro}
\begin{SCfigure}[0.95][h]
\centering
\caption{\textbf{The global workspace (GW) topology for "RNNs of RNNs".} Here we contrast an exemplary depiction of a random connectivity structure between subnetworks in a network of networks (top) versus a global workspace connectivity structure (bottom). Between subnetwork connectivity can be abstractly depicted with an adjacency matrix $\mathbf{A}$ of size $p \times p$ for a network with $p$ subnets. Yellow arrows exist in this depiction between subnetworks $i$ and $j$ for all entries $A_{ij} = 1$. Using the convention of \citet{kozachkov2022rnns} in practice, we will define the weights within subnetworks (white arrows) across the multi-area network via the block diagonal matrix $\mathbf{W}$, and then define the between area weights (grouped as yellow arrows) via a second matrix $\mathbf{L}$ with block diagonal zero. In the GW network (bottom), there is one central subnetwork connected to all other subnetworks, but no other inter-area weights. For a global workspace model with GW subnet containing $g$ units, $\mathbf{L}$ will have $g$ non-zero rows and $g$ non-zero columns.}
\includegraphics[width=0.45\textwidth,keepaspectratio]{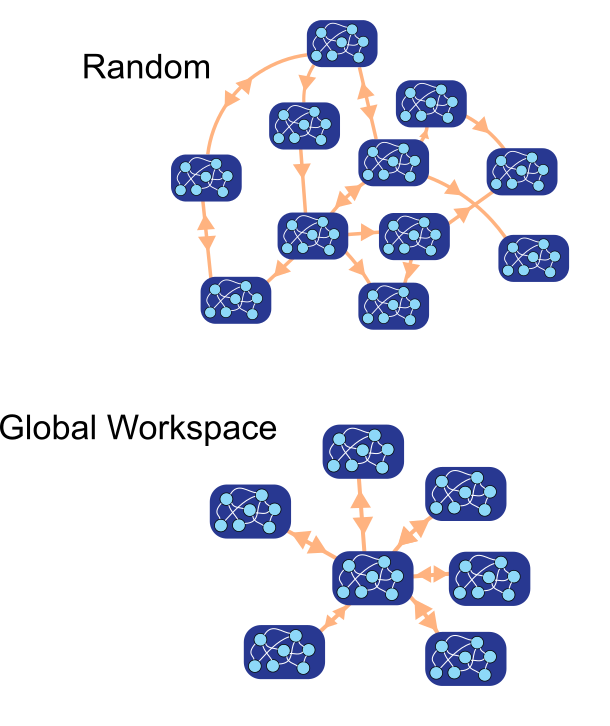}
\label{figure:gw-summary}
\end{SCfigure}

Besides studying modular structures that result from different training paradigms, we can take greater inspiration from known neuroanatomical properties to impose a specific multi-area topology on an "RNN of RNNs", thereby investigating how a given structural constraint impacts various facets of model performance. One such structure of special salience is derived from the global workspace (GW) theories of cognition \citep{newell1972human,baars1993cognitive,dehaene1998neuronal,butlin2023consciousness,volzhenin2022multilevel}. The GW combination is composed of a central subnetwork through which all other subnetworks interact (Figure \ref{figure:gw-summary}). The core idea of the global workspace is that it serves as a shared latent space for communication across the network. It is a prefrontal cortex-like structure that in a human cognitive interpretation represents conscious experience and sensory imagination. 

Because there have been recent calls from those at the intersection of computational neuroscience and deep learning to leverage GW theory in design of network architectures \citep{vanrullen2021deep}, it is convenient that a GW structure can easily be implemented within the established "stable RNNs of RNNs" model class \citep{kozachkov2022rnns}. In addition to the global workspace themes arising within discussion of emerging multi-area RNN research in computational neuroscience \citep{yang2021towards,vanrullen2021deep}, \citet{goyal2021coordination} recently utilized the concept of a shared global workspace in the context of Transformer attention mechanisms. While their work demonstrated great success and brought a spotlight to GW ideas in machine learning, their model does not have any dynamics and thus does not address questions relating to recurrent modular architectures. 

\FloatBarrier

\subsection{Stability in Theory and Practice}
\label{subsec:stability-intro}
A complication of recurrence is that it enables instability, an undesirable property for engineered systems. In a neural network model, instability can prevent learning of certain sequences, increase susceptibility to noise, and decrease behavior interpretability, among other pitfalls \citep{slotine1991applied}. In a biology context, stability can be motivated by the fact that neural firing rates are reproducible across trials with different initial conditions and noise levels \citep{Churchland2010}. Indeed, stability is a central component in several influential neuroscience theories \citep{hopfield1982neural,seung1996brain,murphy2009balanced}. Furthermore, both excitatory/inhibitory (E/I) balance and synaptic sparsity are closely related to stability of dynamical systems models \citep{kozachkov2020achieving}. Empirically, E/I imbalance has been observed in relevant disease states, such as the hyperexcitability consistently found in epilepsy \citep{Ziburkus2013}. Sparsity is an especially common recurring theme in neurobiology, with only $\sim 0.00001\%$ of potential connections forming in the human brain \citep{slotine2012links}. Like E/I balance, synaptic sparsity has been linked to disease states, with excessive synaptic pruning believed to occur in schizophrenia and, inversely, insufficient pruning in autism \citep{Stevens,Sakai}.  

One of the largest gaps in stability theory for RNNs at present reflects the broader paucity of theoretical neuroscience work on multi-area recurrent networks \citep{kozachkov2022rnns}.  Loosely speaking, a dynamical system is said to be globally contracting if given any two system trajectories, they will converge to each other exponentially in a valid distance metric, regardless of initial conditions \citep{lohmiller1998contraction}. A major benefit of contractive stability is that there are a number of ways in which contracting systems can be combined whilst preserving stability of the overall system. Although the importance of building modular systems is clear, in general there is no guarantee that a combination of stable systems will itself be stable. Because contraction analysis tools allow complicated contracting systems to be built up recursively from simpler elements, this form of stability is well suited for neural and other biological systems \citep{slotine2012links}. Moreover, contracting combinations can be made between systems with very different dynamics, as long as those dynamics are each contracting -- thereby facilitating a great deal of flexibility in constructing stable modular systems \citep{slotine2001combos,modular}. 

Note that sparsity can be of heightened interest in the "RNNs of RNNs" framework because it encourages individual subnetworks to take on truly unique topologies, thus providing clearly distinguishable module types to use in building combinations. Recent neurobiological evidence suggests that great diversity in the number and spatial arrangement of thalamocortical synapses across neurons is what enables strong information processing capabilities despite the extreme level of sparsity, and that this confers benefits in robustness and flexibility \citep{Balcioglu2023} -- properties closely linked to stability analysis. Additionally, \citet{Beyeler2019} point out that the degree to which neural activity matches a sparse representation can vary by brain region, yet another argument for greater study of multi-area networks through a control theory lens. In the machine learning literature, there is even a gap in this topic on the individual RNN level, as there has only recently been a reemergence of works focused on sparsity in the weights of modern trainable RNN models \citep{Khona2022}. That is likely a result of the practical success of artificial neural networks over the last decade, which have in large part focused on fully connected models. It is unsurprising from a neurobiological perspective though that an RNN with the assumption of a fully connected weight matrix would be infeasible to scale, and RNNs have indeed lost traction recently due to limitations on their size compared to other models \citep{transformers}.

\section{The "Stable RNN of RNNs" Model Class}
\label{sec:model-intro}
Throughout this paper we use a multi-area recurrent neural network model, i.e. an "RNN of RNNs", as introduced with stability conditions in \citep{kozachkov2022rnns}. To summarize, each individual RNN used as a subnetwork is described by a well-established continuous-time nonlinear RNN model equation \citep{Wilson_Cowan_1972}, and then the units within those different individual RNNs are connected across the subnetwork "areas" via a different component of the overall model. To construct \emph{stable} networks of networks, we perform these inter-area combinations in a way that keeps the overall system dynamics stable so long as the subnetwork dynamics are independently stable, via contraction analysis \citep{modular}. Unfortunately, there is minimal prior work on stability of "RNNs of RNNs", so we focus our mathematical background on explaining previous results \citep{kozachkov2022rnns}. For additional discussion of stability conditions for \emph{individual} RNNs, see the introduction in \citep{kozachkov2022rnns}.

\subsection{Model Definition}
Consider a collection of $p > 2$ nonlinear RNNs, where the $i^{th}$ RNN has $n_i$ units and evolves according to the standard continuous-time RNN equation:
\begin{equation}\label{eq:RNN}\tag{individual nonlinear RNN dynamics}
\dt{x}_i = -x_i + W_i \phi(x_i) + u_i(t) \ \in \fR^{n_i} 
\end{equation}
The nonlinearity $\phi$ is assumed to be slope-restricted so that there exists some $g_{\phi} \geq 0$ such that $0 \leq \phi' \leq g_{\phi}$, but is otherwise unconstrained. Note that \eqref{eq:RNN} describes the evolution of an $n_i$-dimensional state \textit{vector} here. That is, the index $i$ refers to the index of an entire network in the collection, not of an individual neuron in the RNN. While each of the $p$ RNNs could theoretically be studied independently, here we will use these RNNs as \textbf{subnetworks} of a larger multi-area RNN system. \\

\noindent To describe the RNN of RNNs, we stack the state vectors $x_{i}$ of the $p$ subsystems into a larger state vector $\x$, of dimension $N = \sum_{i = 1}^p n_i$. Similarly, we stack the weight matrices $W_{i}$ into a block diagonal matrix $\mathbf{W}$ of dimension $N \times N$. Thus we have the following to describe the combined dynamical system, and we define the input vector $\*u$ similarly:
\[\x \equiv \begin{bmatrix} x_1 \\ \vdots \\ x_p\end{bmatrix} \in \fR^N \hspace{1cm}  \W \equiv \text{BlockDiag}(W_1,\dots,W_p ) \ \in \fR^{N \times N}\]
Of course because $\mathbf{W}$ is block diagonal, each component subnetwork of the combined network would evolve entirely independently of all other subnetworks in this framework. To specify the interactions \textit{between} subnetwork modules, we therefore define the block off-diagonal matrix $\*L \in \fR^{N \times N}$. For now we assume connections between subnetworks are linear as in \citep{kozachkov2022rnns}. In sum, the whole "RNN of RNNs" system has dynamics:
\begin{equation}\label{eq:RNN_of_RNNs}\tag{overall "RNN of RNNs" dynamics}
\dt{\x} = \*f(\x,t) = -\x + \W\phi(\x) + \*u(t) + \*L \x \  \in \fR^{N} 
\end{equation}

\subsection{Established Stability Conditions}
The system with dynamics described by \eqref{eq:RNN_of_RNNs} is not inherently stable. To construct provably stable models in this framework, we use the two-pronged approach introduced in \citep{kozachkov2022rnns}:
\begin{enumerate}[itemsep=0em]
    \item Ensure that each individual nonlinear RNN subnetwork $i$ is independently contracting in some known metric $M_i$. This could be done using many different established contractive stability conditions for the RNN \eqref{eq:RNN}, including those presented in Theorems 1-5 of \citep{kozachkov2022rnns}. Then define the stacked metric matrix for the combined system as $\mathbf{M} \equiv \text{BlockDiag}(M_1,\dots,M_p ) \ \in \fR^{N \times N}$.
    \item Ensure that these contracting RNN subnetworks are connected to each other via $\mathbf{L}$ in a manner consistent with established contraction-preserving combination properties in the metric $\mathbf{M}$. One such methodology would be the negative feedback parameterization for $\mathbf{L}$ derived by \citep{kozachkov2022rnns}, which we make extensive use of in our empirical work here. 
\end{enumerate}
\noindent We will report on new stability conditions for the model \eqref{eq:RNN_of_RNNs} in section \ref{sec:math-results}, but as our experimental results (section \ref{sec:code-results}) utilize a few conditions from \citep{kozachkov2022rnns}, we first elaborate on that work.

\paragraph{Restricted weight magnitudes in the nonlinear RNN subnetworks.}
Theorem 1 of \citep{kozachkov2022rnns} was used to implement the "Sparse Combo Net" stable multi-area architecture, which is the state of the art (SOTA) for sequence processing benchmarks amongst provably stable RNNs at present. That theorem provides a sufficient condition for contraction of an \emph{individual} nonlinear RNN $i$ in terms of a constraint on its internal weight matrix $W_{i}$. If the weight magnitude matrix $|W_{i}|$ obtained via the elementwise absolute value is linearly stable (i.e. when $\phi(x) = x$), then the nonlinear system \eqref{eq:RNN} must be contracting, and will actually be contracting in any diagonal metric that the linear system with $|W_{i}|$ is. The primary principle underlying their proof is the fact that linear stability is equivalent to diagonal stability for Metzler (i.e. nonnegative off-diagonal) matrices \citep{narendra2010metzler}. The condition is very simple to check because of its deep link to linear stability, and it is satisfied by a number of matrices with interesting sparse structures. Moreover, it is highly useful for the aforementioned two-pronged approach that a contraction metric can be easily computed for any nonlinear system \eqref{eq:RNN} satisfying this condition, and it is also convenient that a diagonal $M_{i}$ can always be found. 

Because inhibitory weights cannot contribute to satisfying this restricted weight magnitudes condition, nonlinear RNNs clearly must do so through weaker or fewer weights. As the number of units in an RNN increases, it becomes less and less practical to meet the condition by keeping individual weight magnitudes in the dense matrix small; that would require exceedingly tiny values thereby impacting the network's ability to learn effectively. However, a sparsely connected network could have strong non-zero weights while remaining globally contracting per Theorem 1 of \citep{kozachkov2022rnns}. Empirically, \citet{kozachkov2022rnns} leveraged sparsity in the subnetworks of their aptly-named Sparse Combo Nets to achieve the strong performance that they demonstrated for their stable multi-area architectures. Both the work in \citep{kozachkov2022rnns} and the results we will present here are thus deeply linked with sparsity, which is an important property found in neurobiological systems.

\paragraph{Negative feedback parameterization.}
With a multi-area approach, it is possible to train the connections \emph{between} subnetworks, i.e. $\mathbf{L}$ in \eqref{eq:RNN_of_RNNs}, on a particular task without updating the weights \emph{within} each subnetwork, i.e. $\mathbf{W}$ in \eqref{eq:RNN_of_RNNs}. Such a learning algorithm has close ties to a number of the themes found in biological modularity, like the facilitated variation theory of evolution \citep{gerhart2007fv}. This general idea may also be considered as an extension of the historical echo state network idea \citep{jaeger2001echo} to a more complex "systems of systems" context. Additionally, it is the methodology used to achieve SOTA performance with the aforementioned Sparse Combo Net in \citep{kozachkov2022rnns}. There, nonlinear RNN subnetworks were simply randomly initialized using size, sparsity, and entry magnitude settings, and only those meeting the desired weight magnitudes condition were kept. Those subnetworks then had their \emph{intra-}area weights fixed throughout training; only the linear \emph{inter-}area weights were optimized within the "RNN of RNNs". 

Because our empirical work (section \ref{sec:code-results}) will involve the Sparse Combo Net, it is important to understand how negative feedback connections in this architecture are learned, to optimize performance whilst maintaining a contraction guarantee for the overall system. To do so, we refer to the parameterization for $\mathbf{L}$ in \citep{kozachkov2022rnns}:
\begin{equation}\label{eq:comboRNN}\tag{linear negative feedback parameterization}
\mathbf{L} \equiv \mathbf{B} - \mathbf{M}^{-1}\mathbf{B}^T\mathbf{M}
\end{equation}
where $\mathbf{M}$ is the stacked matrix of contraction metrics for each of the nonlinear subnetwork RNNs $i$ described by \eqref{eq:RNN}. The core idea behind the parameterization (\hspace{-1.25mm}\autoref{eq:comboRNN}) is that when perceived \emph{in the appropriate metric} $\mathbf{M}$, the connections of $\mathbf{L}$ boil down to pure antisymmetric negative feedback. Indeed, when $\mathbf{M} = \mathbf{I}$, $\mathbf{L} = -\mathbf{L}^{T}$. Mathematically, $\mathbf{B}$ can be an arbitrary square matrix, and this parameterization of $\mathbf{L}$ will guarantee contraction for the overall system \eqref{eq:RNN_of_RNNs}. As such, $\mathbf{B}$ can be treated as a trainable tensor in a deep learning framework. To preserve the interpretation of $\mathbf{L}$ as specifying only \emph{inter-}area weights however, $\mathbf{B}$ should be constrained to be block off-diagonal. The PyTorch implementation of Sparse Combo Net indeed masks $\mathbf{B}$ in this way, and also masks it to be lower triangular so that the number of trainable parameters matches the actual degrees of freedom for the negative feedback relationship used. 

\section{Mathematical Results}
\label{sec:math-results}
While important theoretical foundation was laid for the stable "RNNs of RNNs" model class in \citep{kozachkov2022rnns}, there were a number of limitations to the presented stability conditions. The focus on global contraction, and moreover global contraction in a constant (and often diagonal) metric, is a much stricter notion of stability than what we expect of the biological brain. Overall, \citet{kozachkov2022rnns} made minimal assumptions about the model, for example allowing all-to-all inter-areal connectivity. Although it is useful to have generally applicable theorems, this further increased the restrictiveness of the conditions that were feasible to prove. By assuming additional constraints that are of prior neuroscientific or computational interest, we can more readily find stability conditions with increased permissiveness for salient special cases. 

One category of assumption that warrants further consideration is the restriction of connections between subnetworks. Not only does allowing for a fully connected modular structure stray far from neurobiological inspirations, but theoretically a great deal can be said about network dynamics of different sparse models based purely on the underlying graph structure -- including specification of constraints on stable fixed points \citep{Curto2023}. Study of graph structure in multi-area RNNs has high potential, but was a topic hardly covered in \citep{kozachkov2022rnns}. 

A major limitation of the initial "RNNs of RNNs" model, introduced by the requirement for generally provable stability \citep{kozachkov2022rnns}, was the (lack of) activation function on the \emph{inter-}area connections in (\hspace{-1.25mm}\autoref{eq:RNN_of_RNNs}). Requiring all negative feedback weights to be linear degrades what could be meaningfully gained from recursive composition of networks using the contraction combination properties. Furthermore, the \emph{inter-}area connections trained by \citep{kozachkov2022rnns} were constrained to represent pure negative feedback, yet another restriction on the possible expressivity of their model. 

In sum, because of the wide ranging opportunities for expansion on previous theoretical results \citep{kozachkov2022rnns}, our mathematical results largely involve relaxation of certain stability conditions originally reported there. By considering constraints on inter-area topologies and inter-area weight magnitudes instead, we avoid the restrictiveness of linear pure negative feedback. Specifically, we prove a more permissive multi-area network contraction condition in the special case of a \textbf{global workspace} modular topology (subsection \ref{subsec:gw-math}). We also close with mathematical remarks linked to our current experimental aims on the discovery of effective sparse multi-area connection topologies (subsection \ref{subsec:experiment-context}). 


\subsection{Global Workspace Combinations}
\label{subsec:gw-math}
Here we introduce a novel contraction-preserving combination mechanism termed the Global Workspace Combination, drawing inspiration from the global workspace (GW) theories of cognition detailed above (section \ref{subsubsec:gw-intro}). The GW combination is composed of a central contracting component that interacts with other contracting elements through separate feedback mechanisms. While a GW-inspired structure could be imposed within the existing framework of \cite{kozachkov2022rnns} -- and that is something we will explore empirically (section \ref{subsec:negative-feedback}) -- we also leverage some of the unique advantages of this topology to derive new, relaxed stability conditions specific to GW combination. Notably, we can utilize our GW-specific procedure for building stable networks of networks in a framework that does allow for nonlinear connections between modules. Thus we now consider the following adjusted model for the dynamics of our multi-area RNN:  
\begin{equation}\label{eq:RNN_of_RNNs_nonlinear}\tag{global workspace "RNN of RNNs" dynamics}
\dt{\x} = \*f(\x,t) = -\x + \W\phi(\x) + \*u(t) + \*L\psi(\x) \  \in \fR^{N} 
\end{equation}
Where $\psi$ is an inter-areal nonlinearity meeting the same assumptions as those required for $\phi$, i.e. $\psi$ is slope-restricted such that $0 \leq \psi' \leq g_{\psi}$. There are no other distinctions between this system and that of \eqref{eq:RNN_of_RNNs} introduced in section \ref{sec:model-intro}.

For the architecture constrained to have GW topology, the inter-area connectivity matrix $\*L$ takes on a special structure, which can be exploited in proofs of stability and learning convergence. We adopt the convention that the `central' subnetwork (Figure \ref{figure:gw-summary}) is denoted by index 0. Then the $i,j$ block of matrix $\*L$ ($L_{ij}$, corresponding to the weights leaving subnetwork $j$ and arriving at subnetwork $i$) are zero unless $i = 0$ or $j = 0$. Note that the diagonal block $i = 0, \ j = 0$ is zero by definition, because $\*L$ is restricted to only connections \emph{between} distinct subnetworks. Given this global workspace structure for $\mathbf{L}$, along with subnetworks known to be individually contracting in some identifiable metric, we now set out to find stability conditions for the resulting system with dynamics that evolve according to \eqref{eq:RNN_of_RNNs_nonlinear}. We can also think of this combination property as utilizing repeated nonlinear couplings, to ultimately create a flower-like structure.  \\

\noindent Applying contraction analysis as introduced by \citet{lohmiller1998contraction}, we begin with the symmetric part of the Jacobian for the system (\hspace{-1.25mm}\autoref{eq:RNN_of_RNNs_nonlinear}):
\[\frac{1}{2} \big[\*J + \*J^T \big] = \underbrace{-\*I + \frac{1}{2} \big[\*W\*D_{\phi} + \*D_{\phi}\*W^T \big]}_{\text{Intra-areal Jacobian}} + \underbrace{\frac{1}{2} \big[ \*L\*D_{\psi} + \*D_{\psi}\*L^T \big]}_{\text{Inter-areal Jacobian}}\]
For convenience, we assume that all $p$ subnetworks are independently contracting in the identity metric with rate $\lambda$. We then show that a sufficient condition for overall system contraction in the identity metric may be given in terms of the spectral norms of the individual \emph{inter}-area weight matrices:
\[\boxed{||L_{ij}|| \leq \frac{\lambda}{g_{\psi}\sqrt{p-1}}}\]
Note that this condition will be more easily met by $\mathbf{L}$ when the subnetworks defined in $\mathbf{W}$ are more strongly contracting. It will also be more easily met by an $\mathbf{L}$ that includes some amount of sparsity among the connections that form to/from the global workspace for each subnetwork. By contrast, the negative feedback connections parameterized in \citep{kozachkov2022rnns} assume by default that if two subnetworks are joined in negative feedback, their units will be fully connected to each other. Aside from the ability to have nonlinear weights between areas, the new GW combination contraction condition has the additional benefit of allowing for some subnetworks to form stronger connections going into the global workspace and other subnetworks to form stronger connections coming out of the global workspace. 

\paragraph{Proof.} If all subnetworks are contracting in the identity metric, then the intra-areal part of the Jacobian above must be negative definite, so that the overall equation can be upper-bounded as follows:
\begin{equation*}
\frac{1}{2} \big[\*J + \*J^T \big] \preceq -\lambda\*I + \frac{1}{2} \big[ \*L\*D_{\psi} + \*D_{\psi}\*L^T \big] = -\lambda\*I + (\*L\*D_{\psi})_{sym}
\end{equation*}
Where $\mathbf{D}_{\psi}$ is a time-varying positive semi-definite diagonal matrix representing the derivative of the activation function $\psi$ when evaluated at the present state for each unit. The maximum value for each entry of $\mathbf{D}_{\psi}$ is then $g_{\psi}$. Because \citet{kozachkov2022rnns} used only linear inter-area weights, $\mathbf{D}_{\psi} = \mathbf{I}$ in all models from \citep{kozachkov2022rnns}. As mentioned, when $\mathbf{M} = \mathbf{I}$, the negative feedback parameterization used in \citep{kozachkov2022rnns} resulted in $\mathbf{L} = -\mathbf{L}^T$ so that $\mathbf{L}_{sym} = \mathbf{0}$. Here we do not have $\mathbf{D}_{\psi} = \mathbf{I}$, and we also do not wish to make our inter-area weights constrained into perfect negative feedback, so we must find other ways to uniformly upper bound the maximum eigenvalue of $(\*L\*D_{\psi})_{sym}$ as needed. We will consider properties of the block matrices contained within $\mathbf{L}$ in order to apply the Schur Complement Theorem, taking advantage of the structural constraints imposed by GW combination to complete the proof. \\

\noindent Recall that $\mathbf{L}$ has block diagonals equal to 0, and so will $(\*L\*D_{\psi})_{sym}$. The off-diagonal block terms of $(\*L\*D_{\psi})_{sym}$ may be written as:
\[ \frac{1}{2} \big[ L_{ij} D_{\psi_{j}} + D_{\psi_{i}} L_{ji}^T \big] \ \ \in \fR^{n_i \times n_j} \]
For subnetworks $i$ and $j$ with $n_i$ and $n_j$ units respectively. In the case of GW combination, we need only worry about the off-diagonal blocks with $i=0$ or $j=0$, as all others are definitionally zero. For $i = 1$ to $p$, we consider the following blocks:
\[ \frac{1}{2} \big[ L_{i0} D_{\psi_{0}} + D_{\psi_{i}} L_{0i}^T \big] \]
Then for $j = 1$ to $p$, we consider the following blocks, which are each the transpose of the corresponding $i$ block:
\[ \frac{1}{2} \big[ L_{0j} D_{\psi_{j}} + D_{\psi_{0}} L_{j0}^T \big] \]
We are interested in using the spectral norm of each such block. Because the spectral norm of a matrix and its transpose are equivalent, this reduces our consideration to the max eigenvalues for $p-1$ total unique matrices, defined as the symmetric part of each above off-diagonal block. We thus have for $i = 1$ to $p$:
\[ G_{i} = \frac{1}{4} \big[ L_{i0} D_{\psi_{0}} + D_{\psi_{0}} L_{i0}^T + L_{0i} D_{\psi_{i}} + D_{\psi_{i}} L_{0i}^T \big] \]
We can then note that the spectral norm of $G_{i}$ (which equals the spectral norm of the $0,i$ and $i,0$ off-diagonal blocks of $(\*L\*D_{\psi})_{sym}$) can be upper bounded in terms of the spectral norms of $L_{i0}$ and $L_{0i}$. In particular if we assume that:
\[||L_{i0}|| \leq \ell \hspace{1cm} \text{i.e.}\hspace{1cm} \frac{L_{i0} + L_{i0}^T}{2} \preceq \ell \]
\[(\text{where} \preceq \text{denotes an inequality on the maximum eigenvalue}) \] 
and
\[||L_{0i}|| \leq \ell \]
Then we have:
\[||G_{i}|| \leq g_{\psi}\ell \]
Using the GW block structure imposed on $(\*L\*D_{\psi})_{sym}$ and the bounds on the spectral norms of all relevant blocks $||G_{i}||$, we can now ensure that the symmetric part of the overall system Jacobian is negative semi-definite via the Schur Complement Theorem \citep{gallier2020schur}. This yields the following sufficient condition for contraction of the overall system in terms of $\ell$, the small-gain criterion for the inter-area weights as desired:
\[\ell \leq \frac{\lambda}{g_{\psi}\sqrt{p-1}}\] 

\begin{remark}
    The GW combination property can be used in a similar way to negative feedback as a primitive for recursive construction of "RNNs of RNNs of RNNs (...)". This enables a number of interesting structures to be built at scale, including e.g. a GW combination of multiple all-to-all negative feedback multi-area networks.
\end{remark}

\begin{remark}
The above result may be straightforwardly generalized to the case of $\mathbf{M}\neq \mathbf{I}$. Diagonal metrics are particularly simple, and appear naturally when considering hierarchical and skew-symmetric (up to a scalar gain factor) combinations of contracting systems.
\end{remark}

\begin{remark}
Note that in the special case where the inter-area connections are linear (i.e. $\psi(x) = x$), the constraint on $||L_{ij}||$ need not scale with the number of subnetworks $p$. This is because for linear connections $\mathbf{D_{\psi}} = \mathbf{I}$, so skew-symmetry (with respect to a metric) leads to perfect cancellation and thus does not depend on $p$. One could also consider a model with two different types of inter-areal connections between subnetworks, one linear and one nonlinear -- then only the nonlinear connectivity matrix should scale with the number of subnetworks $p$. 
\end{remark}

\subsection{Theoretical Context for Experimental Aims}
\label{subsec:experiment-context}
Our pilot empirical work to be presented in the subsequent section primarily utilizes the model (\hspace{-1.25mm}\autoref{eq:RNN_of_RNNs}) with nonlinear subnetwork stability enforced by the weight magnitudes condition in Theorem 1 of \citep{kozachkov2022rnns} and overall model stability enforced via the condition (\hspace{-1.25mm}\autoref{eq:comboRNN}). However previous work primarily assumes negative feedback will be perfect and all-to-all, there are a few important points to make before exploring variations to the inter-area structure \citep{kozachkov2022rnns}. Particularly in the context of removal of weights or units from the network, as these are topics of high empirical priority, but in general stability preservation cannot be guaranteed with such actions. 

\paragraph{Pruning $\mathbf{L}$.} In the most general case, $\mathbf{L}$ cannot be naively pruned while guaranteeing preserved stability. Even symmetric pruning of $\mathbf{L}$ could potentially disrupt the negative feedback parameterization \eqref{eq:comboRNN}. However when using the weight magnitudes condition to construct $\mathbf{W}$ for the Sparse Combo Net, the metric $\mathbf{M}$ to be used for negative feedback is guaranteed to be diagonal. It is apparent from \eqref{eq:comboRNN} that when $\mathbf{M}$ is diagonal, only changes to the weight $L_{i,j}$ within $\mathbf{L}$ can influence $L_{j,i}$, and if $L_{i,j} = 0$ then we must have $L_{j,i} = 0$. Given this, it is safe to pairwise remove any negative feedback connections between individual units in a Sparse Combo Net. While $\mathbf{B}$ could be pruned in any implementation of the "RNN of RNNs", pruning of the inter-area weights directly is a more natural operation when possible. 

\paragraph{Ablating Units.} Because diagonal stability guarantees that all principal submatrices are also diagonally stable \citep{narendra2010metzler}, removing neuron(s) from a nonlinear subnetwork that is stable via the weight magnitudes condition will always be guaranteed to preserve stability. This is not generally true, though it is highly intuitive here, given that removal of inhibitory units would not adversely impact a magnitudes-based stability condition. Further, the Hurwitz Metzler matrices used to evaluate the Sparse Combo subnetworks will remain contracting in the same metric $\mathbf{M}$ by simply removing the $k$th row and column if neuron $k$ was removed from the network \citep{ennis2023}. In conjunction with the above observation on pruning $\mathbf{L}$ in the Sparse Combo Net model, it becomes apparent that units can be ablated across this multi-area RNN without loss of stability guarantee. 

Enforcing stability preservation upon neuron removal has a number of functional benefits. Robustness to loss of units has obvious safety implications, and furthermore the ability to remove units may also play a role in learning and development. Dropout, a successful regularization technique for deep neural networks, involves randomly removing a fraction of units for each training round \cite{goodfellow2016deep}. Many Neural Architecture Search algorithms include removing units as a potential evolutionary step. Even in biology, a large number of neurons undergo programmed cell death during human development, and this pattern is conserved across many animals \cite{hutchins1998death}. It is likely important not only to be able to combine stable blocks without concern for stability of the overall system, but also to be able to take blocks away without concern. 

\paragraph{Imperfect Negative Feedback.} The parameterization \eqref{eq:comboRNN} leads to $\mathbf{L}$ that connects subnetworks in $100\%$ skew-symmetry up to a metric transformation. This is of course one contraction-preserving way to combine contracting modules, but it is not strictly necessary. A symmetric component could be present in $\mathbf{L}$ as long as it does not "overpower" the negative definite component of $(\mathbf{M}\mathbf{J})_{sym}$ provided by the \emph{intra-}area dynamics. Although it is remarked upon in the theoretical results of \citep{kozachkov2022rnns}, the experimental implementation of negative feedback training still enforces a perfect negative feedback assumption. But signal in $\mathbf{L}_{sym}$ could provide additional expressivity for the network while maintaining a strong stability certificate.


\section{Experimental Results}
\label{sec:code-results}
Broadly, we have a few different categories of major empirical questions about the stable "RNNs of RNNs" model class stemming from previous work \citep{kozachkov2022rnns}:
\begin{itemize}[itemsep=0mm]
    \item \textbf{Inter-area collaboration.} What sort of adjacency structures (e.g. density levels, graph properties) for connections \emph{between} subnetworks will perform best on sequence learning tasks? Can we algorithmically discover such structures for a given task? Can we obtain good performance with connectivity structures instead pre-specified to match task or neuroanatomical priors? Most salient here, how will a global workspace (GW) adjacency perform?
    \item \textbf{Intra-area properties.} How do settings for individual modules contribute to overall network performance? Can mixing different sparsity levels, stability constraints, time constants, activation functions, etc. across the set of modules improve network expressivity? Most salient here, what properties will be most useful for the central subnetwork in a GW model, and how might that vary depending on properties of the other available subnetworks?
    \item \textbf{Interpretability, flexibility, and robustness.} Because both modularity and stability carry a number of potential advantages beyond basic training performance metrics, we will also want to ask questions about the interpretability of different networks, their ability to continually learn, their robustness to adversarial attacks, and so forth.
\end{itemize}
\noindent This paper aims to lay groundwork for research across these many distinct but related questions, especially on the topics of connectivity \emph{between} modules and the use of modular structure to improve model interpretability. We are particularly interested in the performance and other network properties that may result from imposing a global workspace multi-area topology.

\subsection{Methods Summary}
\label{subsec:methods-summary}
Like in \cite{kozachkov2022rnns}, we evaluated different provably stable "network of networks" architectures using sequence learning benchmark tasks. Previous work \citep{kozachkov2022rnns} focused on three sequential image classification tasks often used to evaluate RNN performance: sequential MNIST, permuted sequential MNIST, and sequential CIFAR10. These tasks measure information processing ability over long sequences \citep{le2015simple}, and are particularly common to use in benchmarking provably stable RNNs, as stability in theory has a tradeoff with the ability to integrate information over long timescales -- though in practice stable RNNs have largely performed well with sequence processing \citep{miller2018stable}. For our experiments, we used the 'Sparse Combo Net' code provided by \cite{kozachkov2022rnns}, and performed evaluations using the hardest sequential CIFAR10 (seqCIFAR10) task. Therefore the network's dynamics were defined according to a discrete-time version of equation (\ref{eq:RNN_of_RNNs}), with $\phi =$ ReLU and $\mathbf{L}$ parameterized according to \eqref{eq:comboRNN}. We also retain the '$p \times n$ network' notation of \cite{kozachkov2022rnns} in reporting our results; such a network consists of $p$ distinct subnetwork RNNs, with each such subnetwork RNN containing $n$ units. 

Recall that the Sparse Combo Net does not have the weights within its individual modules trained -- relying only on training the linear (negative feedback constrained) connections between modules, as well as the linear input and output layers. As such, we put our first focus on experimenting with the adjacency structure of the inter-area negative feedback connections. One weakness of the pilot experiments by \cite{kozachkov2022rnns} was that they focused solely on all-to-all negative feedback, which may explain why their architecture failed to scale well beyond size $16 \times 32$. For a $p \times n$ network, we thus define a $p \times p$ adjacency matrix $\mathbf{A}$ that specifies which subnetwork pairings will have inter-area connections between them. $\mathbf{A}$ is constrained to be symmetric and have 0-valued diagonal entries, but otherwise can take on a variety of possible structures. If $A_{i,j} = 0$, then the inter-area weight matrix $\mathbf{L}$ will have $L_{r,c}$ masked to 0 for all $r$,$c$ index pairings where $r$ corresponds to a unit in subnetwork $i$ and $c$ corresponds to a unit in subnetwork $j$.

We otherwise kept network settings very similar to those used by \cite{kozachkov2022rnns}. For additional methodological details, see supplemental section \ref{sec:methods-supp}. Similarly, supplemental results can be found in section \ref{sec:results-supp}.

\subsection{Overview of Empirical Contributions}
We reached $68.91\%$ test accuracy on sequential CIFAR10 during our trials of the stable RNN of RNNs architecture, thereby improving on the state-of-the-art for provably stable RNNs while using a similar number of trainable parameters as prior work (see Tables \ref{table:comp} and \ref{table:sota} for our results and a comparison to existing networks, respectively). Ultimately, we were able to make these improvements by considering multi-area networks with a larger number of modules than \cite{kozachkov2022rnns} but sparser connections between them (section \ref{subsec:negative-feedback}). 

Additionally, we provided proof of concept results for the benefits of the specialized global workspace (GW) adjacency structure for negative feedback, which involves just one "central" subnetwork being connected in negative feedback with all other subnetworks:
\begin{itemize}[itemsep=-0.5mm]
    \item The GW Sparse Combo Net could repeatably perform well on seqCIFAR10 with a very tiny number of trainable parameters (section \ref{subsubsec:gw-results}).
    \item Using a global workspace adjacency structure without the negative feedback constraint entirely failed to learn, underscoring the importance of the theoretical stability conditions (section \ref{subsubsec:gw-results}).
    \item The GW Sparse Combo Net consistently had performance much more resilient to post-training ablation of entire modules than the other multi-area adjacency structures considered. Roles of component subnetworks were also more approachable for interpretation in the global workspace model (section \ref{subsec:rnn-interpretation}).
    \item Increasing the size of the GW subnetwork module was able to improve performance over an otherwise identical multi-area network structure (section \ref{subsec:gw-size-tests}).
\end{itemize}

\subsection{Task Performance Varies with Negative Feedback Structure}
\label{subsec:negative-feedback}
The use of sparsity in subnetworks by \cite{kozachkov2022rnns} to improve Sparse Combo Net performance suggests another direction that could enable better scalability of total network size: enforcing sparsity in the linear feedback weight matrix $\mathbf{L}$. Interestingly, this can be accomplished in meaningfully different ways due to the modular structure of the "RNN of RNNs". To begin investigating the idea of limiting the number of subnetwork modules connected to each other, we tested different feedback sparsity levels in $24 \times 32$ Sparse Combo Nets on the seqCIFAR10 task by randomly sparsifying the adjacency structure between subnetworks in $\mathbf{A}$ whilst maintaining the negative feedback symmetry assumption (Table \ref{table:comp}). 

\begin{table}[h!]
\centering
\begin{tabular}{ | m{8.25cm} || m{1.6cm} | m{1.9cm} | m{1.2cm} | }
\hline
 Name & Trainable Params Count & seqCIFAR10 \newline Test \newline Accuracy & Final \newline Train \newline Loss \\
\hline\hline
10x32 All-to-All Negative Feedback & 50,570 & 62.78\% & 0.9198 \\
\hline\hline
16x32 Global Workspace & 22,538 & 57.92\% & 1.0921 \\
\hline
\rowcolor{LightCyan}
16x32 All-to-All Negative Feedback & 130,058 & 66.85\% & 0.765 \\
\hline\hline
24x32 Random 5\% of Negative Feedback Pairings & 24,074 & 34.23\% & 7.9304 \\
\hline
24x32 Global Workspace & 34,313 & 62.59\% & 0.8161 \\
\hline
\rowcolor{LightCyan}
24x32 Random 50\% of Negative Feedback Pairings & 142,858 & 68.91\% & 0.6062 \\
\hline
24x32 All-to-All Negative Feedback & 293,386 & 53.62\% & 1.3028 \\
\hline\hline
32x32 Global Workspace & 46,090 & 62.5\% & 0.7904 \\
\hline
\rowcolor{LightCyan}
32x32 Random 33.3\% of Negative Feedback Pairings & 183,306 & 68.64\% & 0.541 \\
\hline
\end{tabular}
\caption{\textbf{Comparing seqCIFAR10 performance across different Sparse Combo Net sizes and adjacency structures.} We evaluated different module counts and negative feedback adjacency structures for the Sparse Combo Net using seqCIFAR10. For each trial, the number of trainable parameters in the architecture, the best achieved test accuracy, and the final epoch training loss are reported. Trials with test accuracy exceeding the best overall achieved by \cite{kozachkov2022rnns} (65.72\%) are highlighted. Networks are grouped by the number of subnetworks contained, and ordered within groupings by the number of trainable parameters. Note the "16x32 All-to-All Negative Feedback" network is a repetition of the best performing network of \citep{kozachkov2022rnns} using our slightly modified training function (section \ref{sec:methods-supp}). \newline Overall, there are two major takeaways: we can increase the number of subnetworks and still see some continued gains in performance by utilizing sparsity in negative feedback connections, and we can achieve quite impressive seqCIFAR10 test accuracy with a very small number of trainable parameters by utilizing a global workspace adjacency structure. Compare to other networks tested on these sequence classification benchmarks in Table \ref{table:sota}, reproduced from \cite{kozachkov2022rnns}.}
\label{table:comp}
\end{table}

Sparsifying the negative feedback connections between subnetworks indeed showed promise in mitigating the saturation effect seen with added subnetworks in an all-to-all negative feedback structure (Table \ref{table:comp}). Performance results by feedback adjacency sparsity level took an inverse U shape here, consistent with pilot results reported in the supplement of \citep{kozachkov2022rnns}. Notably, the $24\times32$ Sparse Combo Net with $50\%$ negative feedback (non-zero adjacency connections determined randomly) achieved $68.91\%$ test accuracy on sequential CIFAR10, an improvement of more than $3$ percentage points from the state-of-the-art (SOTA) performance for provably stable RNNs that was established by \citep{kozachkov2022rnns}. This is already a larger increase over SOTA than their original experiments were at the time (Table \ref{table:sota}). Moreover, the number of trainable parameters we consider is similar, as while we have added more subnetworks we are now only training a fraction of the connections between them (Table \ref{table:comp}). By contrast, an all-to-all negative feedback structure for the $24\times32$ Sparse Combo Net performed quite poorly, reaching just $53.62\%$ test accuracy, which was significantly worse than even a $10\times32$ all-to-all Sparse Combo Net, despite having more than $5x$ the number of trainable parameters (Table \ref{table:comp}). Thus our results suggest that Sparse Combo Net is not necessarily limited by scalability, but it cannot scale well if the negative feedback weight matrix remains dense. 

\paragraph{What is scalability?} One might argue that the well-performing larger networks in Table \ref{table:comp} are not actually larger than the best performing networks of \citep{kozachkov2022rnns}, as they retain a similar number of trainable parameters. However, the ability to increase the number of subnetworks or number of units in a sparse multi-area network is an important notion of scalability to consider. Empirically, we have demonstrated that scaling in this way can meaningfully improve performance and (as will be discussed shortly) robustness, without any major methodological changes. Theoretically, such scalability confers additional opportunities outside of the typical training paradigm, including both targeted selection of non-zero adjacency entries via e.g. evolutionary \citep{gaier2019wann} or pruning \citep{frankle2019lth} algorithms, as well as the feasibility of including a large number of modules pretrained on key subtasks or even introducing additional modules post-hoc for e.g. continual learning problems. Ultimately, the graph structure of the network can become another source for optimization, in addition to training of the non-zero weights. Given the strong results observed even with randomly assigned adjacency, there is much reason for optimism along these lines.

\FloatBarrier

\subsubsection{The Global Workspace Network}
\label{subsubsec:gw-results}
Besides computational methodologies for discovering an optimal negative feedback adjacency structure given a task and $\mathbf{W}$, one might also try sparser adjacency structures that are based on prior knowledge, particularly hypotheses from neurobiology. As introduced in section \ref{subsubsec:gw-intro}, a major such network topology comes from the global workspace (GW) theory. The corresponding adjacency structure is quite sparse, non-zero for just one row/column, which maps to the global workspace subnetwork meant to coordinate communication between all other subnetworks. The number of trainable parameters for a GW Sparse Combo Net is therefore very small, but despite this we observed impressive performance on seqCIFAR10 (Table \ref{table:comp}). With nearly $100,000$ fewer trainable parameters than the best performing network of \citep{kozachkov2022rnns}, our $24\times32$ GW Sparse Combo Net achieved $62.59\%$ test accuracy on seqCIFAR10. Only the (questionably stable \citep{kozachkov2022rnns}) Antisymmetric RNN \citep{chang2019antisymmetricrnn} reports a seqCIFAR10 test accuracy with a network using a similarly small number of trainable parameters, and their best performance was just $58.7\%$ (Table \ref{table:sota}). The smallest all-to-all negative feedback Sparse Combo Net we tested was $10\times32$, and while this obtained a very similar test accuracy to the $24\times32$ GW Sparse Combo Net, it still had $\sim 50\%$ more trainable parameters than the global workspace benchmark (Table \ref{table:comp}).  

\paragraph{Repeatability and controls.} Overall, the global workspace adjacency demonstrated impressive test performance given its size, which will enable future experiments well-suited to such a structure, including the coordination of multiple Sparse Combo Nets pretrained on subtasks being combined to train on a larger problem through recursive application of the theoretical principles established by \citep{kozachkov2022rnns}. Crucially for such future applications of the Global Workspace Sparse Combo Net, its overall performance was consistent across repetitions of the same size (Figure \ref{fig:gwt-repeatability}), and the other observed properties (to be discussed) were similar across all trials. Additionally, a $16\times32$ GW Sparse Combo Net with the negative feedback constraint on inter-areal weights ($\mathbf{L}$) removed actually failed to learn entirely, having loss above $10^{23}$ throughout training and final test accuracy at chance. Note that \cite{kozachkov2022rnns} observed very bad performance from an all-to-all $16\times32$ Sparse Combo Net when the negative feedback constraint was removed there, but a possible contributing factor was the $\sim$ doubled number of trainable parameters. For a $16\times32$ GW Net though, this is still only 37898 params, driving home the importance of the stability conferred by the negative feedback constraint. 

\subsection{Leveraging Modular Structure for Interpretability and Resilience}
\label{subsec:rnn-interpretation}
\begin{SCfigure}[0.66][h]
\centering
\includegraphics[width=0.55\textwidth,keepaspectratio]{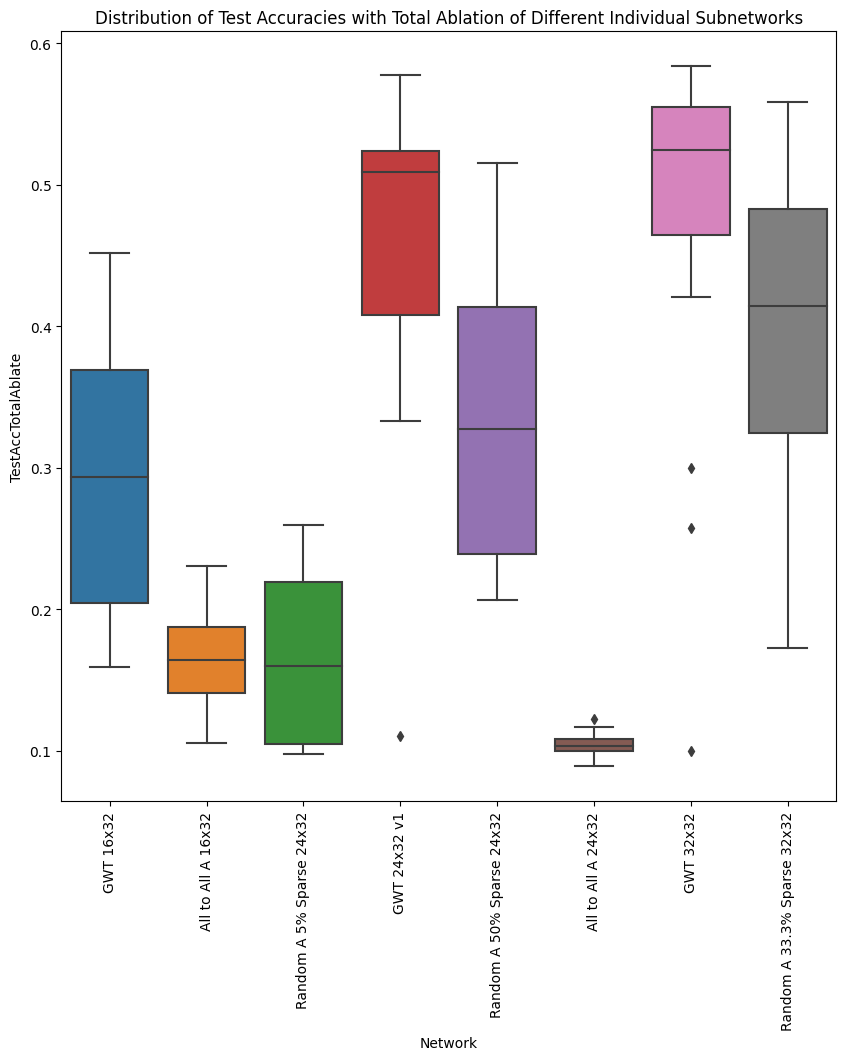}
\caption{\textbf{The effect of post-training subnetwork removal on seqCIFAR10 test accuracy in different stable multi-area network topologies.} To better understand the different Sparse Combo Nets considered, individual subnetworks were ablated at test time to determine impact on performance. The distribution of subnet-ablated test accuracies over the subnetworks contained in that network is depicted for each network here using a box and whisker plot. All trials from Table \ref{table:comp} starting with size $16\times32$ are included, in identical order. Global workspace networks (blue, red, pink) were generally much more resistant to removal of individual subnetworks than other networks of the same size with different connectivity structures were.}
\label{fig:comparing-impact-of-ablation}
\end{SCfigure}

For each subnetwork $i$ in various trained $n \times 32$ Sparse Combo Nets, final test accuracy on sequential CIFAR10 was evaluated with all incoming and outgoing weights to units of subnetwork $i$ fixed at 0 (and no other updates), in order to probe the robustness of the network to loss of modules (Figure \ref{fig:comparing-impact-of-ablation}). This included ablation of all linear input and output weights and hidden to hidden biases associated with that module, in addition to ablating internal nonlinear weights and any linear connections to other subnetworks.  

Unsurprisingly, a larger number of subnetworks typically conferred greater resistance to single subnetwork ablation, and removing the central node in any global workspace model decimated its performance (Figure \ref{fig:comparing-impact-of-ablation}). However global workspace models were in general much more resistant to the effects of subnetwork ablation than other configurations, with more than half of the $24\times32$ GW subnet ablations tested retaining task accuracy above $50\%$. The ablation experiments also further demonstrated the weaknesses of the full negative feedback $24\times32$ Sparse Combo Net, which was notably more impacted by subnetwork ablation than the matched size $50\%$ sparse $\mathbf{A}$ model or even the matched size $5\%$ sparse $\mathbf{A}$ model (Figure \ref{fig:comparing-impact-of-ablation}). Note that the all-to-all $24\times32$ Net did not demonstrate signs of overfitting relative to other models, just generally low performance (Table \ref{table:comp}) -- so the increased susceptibility to module removal helps make more clear its fundamental failings.

\begin{figure}[h]
\centering
\includegraphics[width=0.85\textwidth,keepaspectratio]{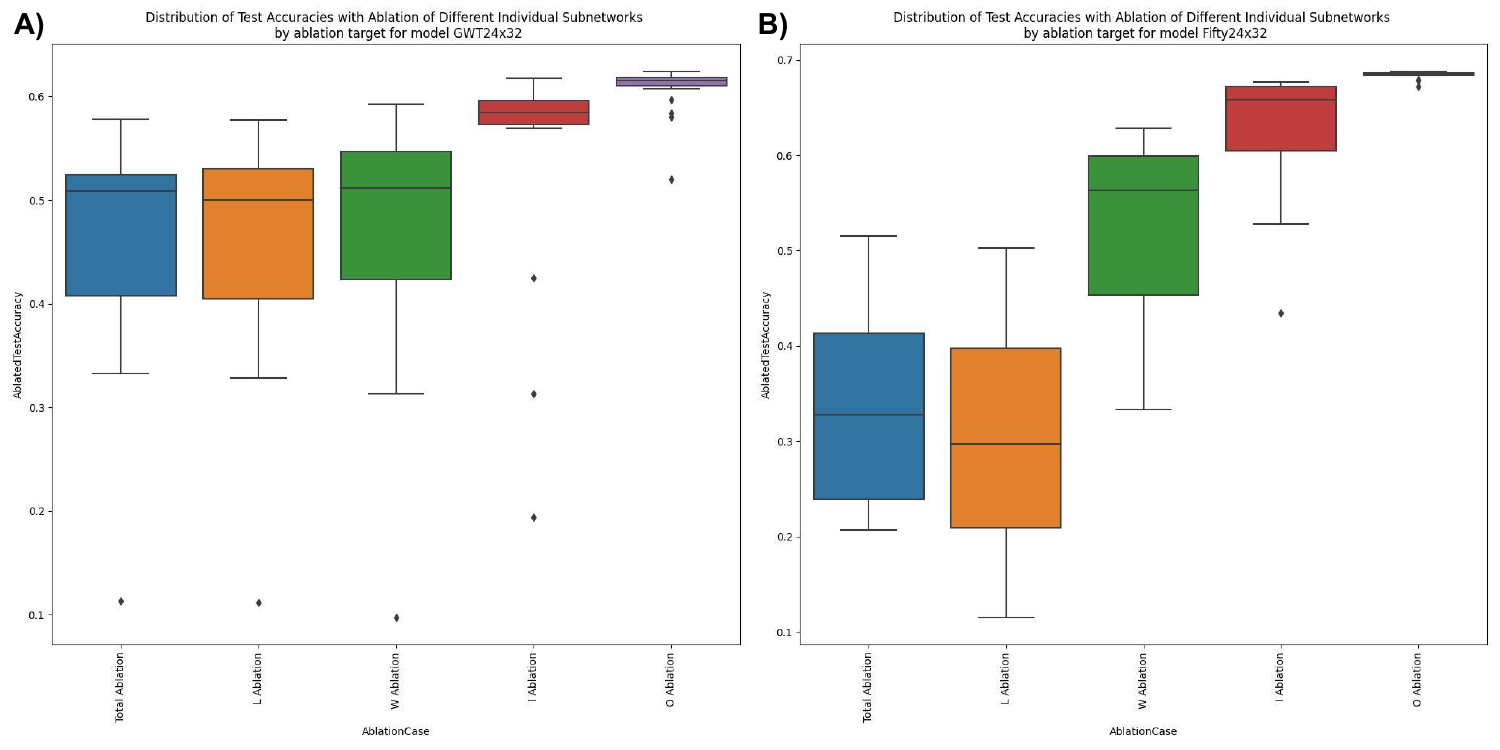}
\caption{\textbf{Targeted ablation by type of weight can further elucidate individual subnetwork roles.} The distribution of subnet-ablated test accuracies over the subnetworks contained in that network is depicted for different weight matrix-specific ablations using a box and whisker plot, for both the $24\times32$ Global Workspace Sparse Combo Net (A) and $50\%$ negative feedback $24\times32$ Sparse Combo Net (B). As a comparison point, total ablation (blue) will match the respective networks' presentations in Figure \ref{fig:comparing-impact-of-ablation}. The isolated weight matrix ablations are then as follows, in order: removal of subnetwork-associated negative feedback weights $\mathbf{L}$ only (orange), removal of subnetwork-associated internal nonlinear weights $\mathbf{W}$ only (green), removal of subnetwork-associated image input weights only (red), and removal of subnetwork-associated prediction output weights only (purple). \emph{Inter-}area connections (orange) were especially important to performance for the $50\%$ $\mathbf{A}$ Sparse Combo Net (B).}
\label{fig:ablation-by-type}
\end{figure}

By next performing subnetwork ablation in a weight matrix specific manner, we demonstrate more concretely how image input weights, nonlinear internal subnetwork weights, negative feedback inter-area weights, and final output weights play different roles in the Sparse Combo Net architecture. It is worth noting that GW networks tended to have similar ablation performance whether entire subnetworks were removed, only inter-area connections were removed, or only intra-area connections were removed (Figure \ref{fig:ablation-by-type}A), while the more connected Sparse Combo Nets were especially hurt by the ablation of the inter-area negative feedback weights (Figure \ref{fig:ablation-by-type}B). More generally, a handful of subnetworks tended to be uniquely important to processing inputs, but output predictions were by and large distributed enough that the final state of any individual subnetwork was unimportant to overall test performance (Figure \ref{fig:ablation-by-type}). 

\begin{figure}[h]
\centering
\includegraphics[width=0.95\textwidth,keepaspectratio]{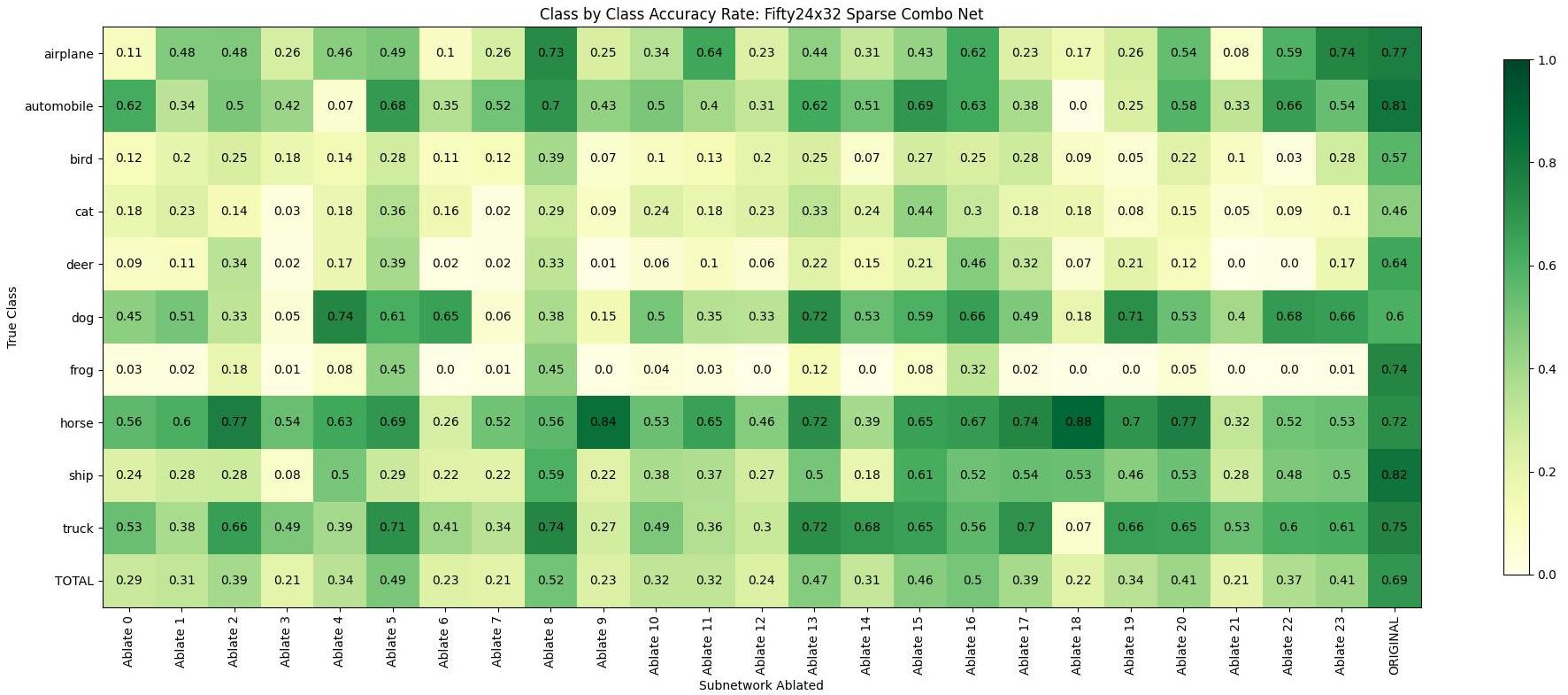}
\caption{\textbf{Class-specific effects of total subnetwork ablation in the $50\%$ sparse negative feedback $24\times32$ stable multi-area network.} Using the same figure generation scheme as Figure \ref{fig:class-specific-ablation-impact}, here we demonstrate the class-specific impacts of \emph{total} subnetwork removal for the different subnets of the $50\%$ $\mathbf{A}$ $24\times32$ Sparse Combo Net. Almost every subnetwork was necessary for the prediction of "frog" in the $50\%$ negative feedback model, and conversely prediction of "horse" did not appear to rely on any individual subnetwork there. A similar pattern emerges even in the less generally disastrous case of removing only internal subnetwork weights (Figure \ref{fig:ablation-fifty-W}).}
\label{fig:ablation-fifty}
\end{figure}

In addition to looking at overall test accuracy results in different ablation cases, we can look at class-specific effects to further elucidate what different roles in network performance each subnetwork may be playing. For the randomly generated moderately sparse negative feedback structures, this did little to separate individual subnets, as many were broadly important (Figure \ref{fig:ablation-fifty}). On the other hand, it did help to identify discrepancies between how the network was predicting different labels. For example, frog and horse had similar accuracy rates with the $50\%$ negative feedback $24\times32$ network fully intact, but the former was severely impacted by removal of nearly any subnetwork while the latter was robust to removal of most subnetworks.  

Systematic analysis of the effects of different weight matrix ablations on different classes revealed distinct subnetwork roles within the $24\times32$ Global Workspace Sparse Combo Net (Figure \ref{fig:class-specific-ablation-impact}), yet another reason this topology is of great interest. For example, we were able to identify specific subnetworks whose final state played an outsized role in prediction of "deer" or of "automobile", other specific subnetworks that were most important for receiving direct image input, and still other subnetworks that had large differences in class by class effects of \emph{inter-} versus \emph{intra-} area weight ablation. Additionally, we found some subnetworks with high importance to a small subset of classes that share some natural commonality -- e.g. subnet 2 of this model was most important to prediction of bird and of airplane (Figure \ref{fig:class-specific-ablation-impact}). In the future, one might consider topologies that purposefully mask image inputs and prediction outputs to only a subset of the modules in the overall network, with input and output modules perhaps kept disjoint. From a global workspace perspective, this could entail restricting the GW subnet to have no direct input or output, which would better reflect the hierarchical structure of sensorimotor processing in the brain. 

\begin{FPfigure}
\centering
\includegraphics[width=0.775\textwidth,keepaspectratio]{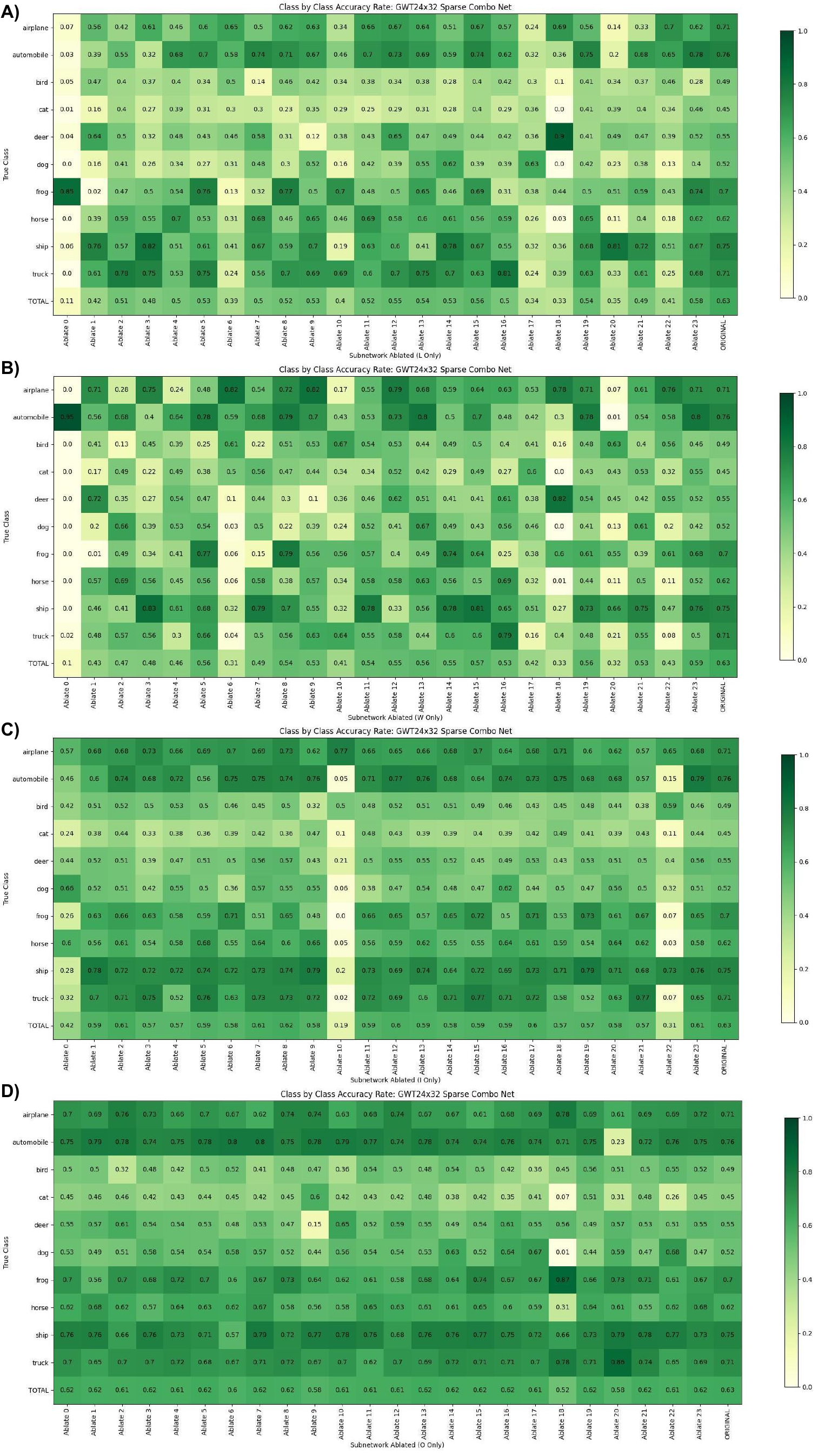}
\caption{\textbf{Class-specific effects of subnetwork ablation with different weight targets for the $24\times32$ Global Workspace Sparse Combo Net.} For each seqCIFAR10 label (and in the last row, across all labels) we report the test accuracy of the $24\times32$ GW Sparse Combo Net on those images when each subnetwork (and in the last column, no subnetwork) had its corresponding weights ablated in the matrix of interest. Cells are colored using the matplotlib 'YlGn' heatmap, from 0.0 to 1.0. Here we systematically focused on removal of only \emph{inter}-area $\mathbf{L}$ weights (A), removal of only \emph{intra}-area $\mathbf{W}$ weights (B), removal of only image input weights (C), and removal of only output prediction weights (D). Although some modules were broadly important, different parts of different subnetworks clearly played specialized roles to an extent. In addition to the global workspace module, subnetworks 10 and 22 were critical to the direct processing of image inputs (C). Furthermore, the class-specific view highlights changes in the type of mistake caused by a particular ablation target. For example, the network was prone to over-prediction of airplane in the absence of external input to subnet 10, yet without the internal weights of 10 the network was especially bad at detecting airplanes (B). While network performance was generally very resilient to loss of output weights from a single subnetwork, there were notable class-specific effects there as well, such as the clear importance of subnetwork 20 final state to the identification of automobiles (D).}
\label{fig:class-specific-ablation-impact}
\end{FPfigure}

\paragraph{In summary...} Because an important potential benefit to modularity is the safe modification to or even removal of individual components \citep{Lorenz2011}, the general robustness of GW models to subnetwork removal (Figure \ref{fig:comparing-impact-of-ablation}) is highly salient. In evolutionary neurobiology, it is hypothesized that modularity of the brain is closely linked with modular task switching \citep{Kashtan2005,Lorenz2011}. In machine learning, we can make parallels with potential ability to perform on continual learning tasks. For example, it is likely that the $50\%$ negative feedback $24\times32$ Sparse Combo Net would quickly forget the concept of frog, as almost every subnetwork was absolutely critical to recognizing any frogs (Figures \ref{fig:ablation-fifty}); the global workspace networks did not demonstrate such "fragilely distributed" concepts by contrast (Figure \ref{fig:class-specific-ablation-impact}).

\FloatBarrier

\subsection{Variations on Subnetwork Components}
\label{subsec:gw-size-tests}
A natural next step from the investigation of modular structures is to the investigation of different component modules. Going forward, this could involve everything from a wider variety of simple initialization settings (size, sparsity level, entry magnitude), to more fundamental differences in subnetwork function (time constant, contraction condition, activation function), and even to use of graph structures with specific aims (subtask pretraining/weight pruning schemes, neurobiologically-inspired constraints like Daleian weights or module duplication and variation via an evolutionary process). Some of these directions might also open additional avenues of experimentation, such as updating of the nonlinear \emph{intra}-subnetwork weights in joint training with the \emph{inter-}area weights of $\mathbf{L}$. In the case of the global workspace architecture we have an obvious starting point for component testing however, through modifying only the properties of the GW module and then observing the resulting network's characteristics. As discussed by \citet{vanrullen2021deep}, the GW could serve best as a translation bottleneck of sorts between the rest of the subnetworks if it were to have higher dimensionality than any one subnetwork, but much lower dimensionality than the sum of the subnetworks. This suggests testing of different sizes for the GW subnetwork would be a useful line of questioning (Table \ref{table:gwt-comp}). 

\begin{table}[h!]
\centering
\begin{tabular}{ | m{8.25cm} || m{1.6cm} | m{1.9cm} | m{1.2cm} | }
\hline
 Name & Trainable Params Count & seqCIFAR10 \newline Test \newline Accuracy & Final \newline Train \newline Loss \\
\hline\hline
16x32 Global Workspace Included & 22,538 & 57.92\% & 1.0921 \\
\hline\hline
24x32 Global Workspace Included & 34,313 & 62.59\% & 0.8161 \\
\hline
24x32 Global Workspace Repetition & 34,313 & 61.52\% & 0.8644 \\
\hline
24x32 Global Workspace Selection & 34,313 & 62.32\% & 0.841 \\
\hline
\rowcolor{Gray}
Size 64 Global Workspace With 24x32 Subnets & 60,810 & 63.72\% & 0.6806 \\
\hline\hline
32x32 Global Workspace & 46,090 & 62.5\% & 0.7904 \\
\hline
\rowcolor{Gray}
Size 128 Global Workspace With 32x32 Subnets & 147,210 & 65.57\% & 0.5532 \\
\hline
\end{tabular}
\caption{\textbf{Performance of Global Workspace Sparse Combo Nets with different central module settings.} The same stats as in Table \ref{table:comp} are presented here, now for an expanded set of global workspace experiments. This includes the initial tests where an otherwise normal node in the Sparse Combo Net was made to be the GW, alongside any repetitions run (Figure \ref{fig:gwt-repeatability}). Additionally, two new trials (highlighted rows) are reported -- a size 64 central module connected to a $24\times32$ set of contributing subnetworks initialized as before, and a size 128 central module connected to a $32\times32$ set of contributing subnetworks initialized as before. In both cases the separate global workspace subnetwork was initialized using a $1\%$ sparsity level and weights between $-5$ and $5$. The larger GW subnetworks both resulted in test accuracy improvements, though they had notably larger size in terms of trainable parameters, and were perhaps slightly overfitted when comparing their training losses to those found in Table \ref{table:comp}.}
\label{table:gwt-comp}
\end{table}

Note that while we used the notation $p \times n$ to refer to different size Sparse Combo Nets through most of this text for convenience, there is of course no requirement for network sizes to be the same, and subnetworks of a variety of sizes can be just as readily described within the existing model definition. Overall, the Size 128 Global Workspace with $32\times32$ Subnets matched the performance of the best network from \citep{kozachkov2022rnns}, and reached this accuracy within the first 200 epochs of training. Both networks had a similar total number of trainable parameters (Tables \ref{table:gwt-comp} and \ref{table:sota}), though the GW network had many more units and thus has the greater potential for expansion to other applications. The size 128 GW network was also quite resistant to post-training subnetwork ablation (Figure \ref{fig:ablation-gwt-large}), unlike the $16\times32$ Sparse Combo Net with all-to-all negative feedback introduced previously \citep{kozachkov2022rnns} was (Figure \ref{fig:comparing-impact-of-ablation}). 

\FloatBarrier

\section{Discussion}

Most work on stability of task-trained RNNs has focused on \textit{single} RNNs. In fact, most work in theoretical neuroscience at large has focused on single RNNs, as have most machine learning applications. But by combining nonlinear control principles with neurobiological inspiration, we can develop stable multi-area RNNs capable of integrating information over long timescales to compete with much larger architectures on benchmark sequence learning tasks. Both the modularity and stability of such networks carry a number of advantages, and there is a massive space of open questions on this class of models -- which if leveraged to their full potential may be the source of a future paradigm shift in deep learning, perhaps akin to the convolutional layer constraints that lead to the great popularity of feedforward neural networks. 

In this paper, we provided proof of concept for improved ability of the Sparse Combo Net \citep{kozachkov2022rnns} at scale when sparsity is introduced in the connectivity structure between subnetworks. We also demonstrated advantages of specialized inter-subnetwork connection topologies, through both new theoretical stability conditions and empirical results using our global workspace inspired stable multi-area network. Critically, the GW combination property introduced here allows for nonlinear connections between subnetworks to be explored while maintaining overall network stability. These results will enable a number of downstream research projects in both machine learning and computational neuroscience. Given the relevance of our work, we close with a sampling of various possible future experimental questions.

\subsection{Future Directions}
\label{subsec:future-dirs}

\paragraph{Immediate empirical follow-ups.} One question introduced above (subsection \ref{subsubsec:gw-results}) was whether we can apply a global workspace negative feedback structure to combine two or more all-to-all negative feedback Sparse Combo Nets after they are initially trained on engineered subtasks e.g. a focus on distinguishing cats and dogs. This has the potential to significantly improve seqCIFAR10 performance given the distribution of typical network mistakes at present (Figure \ref{fig:main-confusion-matrices}), and it would be the first experimental application of the "recursive construction" concept introduced as part of the theoretical model in \citep{kozachkov2022rnns}, as well as the first test of subtask training for addressing a very challenging overall task. It will involve interesting methodological questions too, such as whether the inter-area weights from the first training rounds should be allowed to update further when the larger combination network is trained on the broader task.

Another interrelated set of questions introduced was on the use of pruning or other potential graph learning techniques to identify an optimal adjacency matrix $\mathbf{A}$ for defining inter-area negative feedback connectivity. In addition to evaluating methods for defining $\mathbf{A}$, we can also run more systematic trials involving retraining the same network with only certain properties of interest varied. To date the experiments both here and in \citep{kozachkov2022rnns} have involved entirely reinitializing the networks on each training run with different settings (and repetitions), so performing trials where e.g. $\mathbf{A}$ is varied for training of a network that is otherwise identically initialized could provide important additional evidence for certain hypotheses, as well as relevant practical insights such as easier direct comparison of different pre-training lengths. Furthermore, besides testing pruning algorithms for determining $\mathbf{A}$, we can test pruning algorithms that act directly on $\mathbf{L}$ (in a symmetric manner).

Furthermore, we will investigate combination properties of networks that maintain functionality, rather than merely preserving stability. Such networks could include bespoke recurrent models, such as combinations of winner-take-all (WTA) networks, which are also contracting—i.e., their output dynamics depend on and stabilize with respect to their inputs \citep{rutishauser2011collective}. Ensembles of recurrent neural networks pre-trained on individual tasks may also be considered. Predictive coding, a form of negative feedback, may be employed to learn feedback hierarchies—potentially using RNNs or similar architectures—and can also be analyzed with respect to modular stability \citep{modular,wang2006contraction}. Additionally, we will consider the role of a reward signal in guiding network dynamics and maintaining functional integration.

\paragraph{Utilizing interpretation results.} A distinct category of direct follow-ups to the present results will involve additional tests for interpreting network behaviors, particularly to evaluate hypotheses that may have formed based on our pilot interpretability results (subsection \ref{subsec:rnn-interpretation}). We can track neuron state in response to different inputs to see whether overall activity of different subnetworks in e.g. the $24\times32$ Global Workspace Sparse Combo Net matches expectations for those subnetworks based on the presented class-specific ablation analyses. Additionally, there are questions of repeatability for the observed ablation experiments. If the same networks were retrained with the same initialization values, to what extent would the same specific subnetworks be most susceptible to ablation or clearly associated with the same specific prediction class? Similarly, it could be asked to what extent the overall network can learn to make up for a missing subnetwork if training is resumed without it (or with it rerandomized), rather than restarted. This might speak to early questions about the "critical period" of the architecture.

\paragraph{Opportunities from novel theoretical results.} In section \ref{sec:math-results}, we presented a new relaxed stability condition for the "RNN of RNNs" when a global workspace combination structure is imposed. This condition would allow for a nonlinearity on the \emph{inter}-area weights, making recursive construction of greater interest, and would also enable imbalanced weight magnitudes between connected subnetworks (i.e. not pure negative feedback). However in section \ref{sec:code-results}, we focused on experimentation within the perfect linear negative feedback multi-area architecture. A priority for our future work will thus be to take advantage of our theoretical improvements to ideally make additional strides experimentally.

\pagebreak

\bibliography{biblio}

\pagebreak

\section*{Supplement}

\renewcommand\thefigure{S\arabic{figure}}    
\setcounter{figure}{0}  
\renewcommand\thetable{S\arabic{table}}    
\setcounter{table}{0}  
\renewcommand\thesection{S.\arabic{section}} 
\setcounter{section}{0}

\section{Previous "Sparse Combo Net" Results}
\begin{figure}[h]
\centering
\includegraphics[width=\textwidth,keepaspectratio]{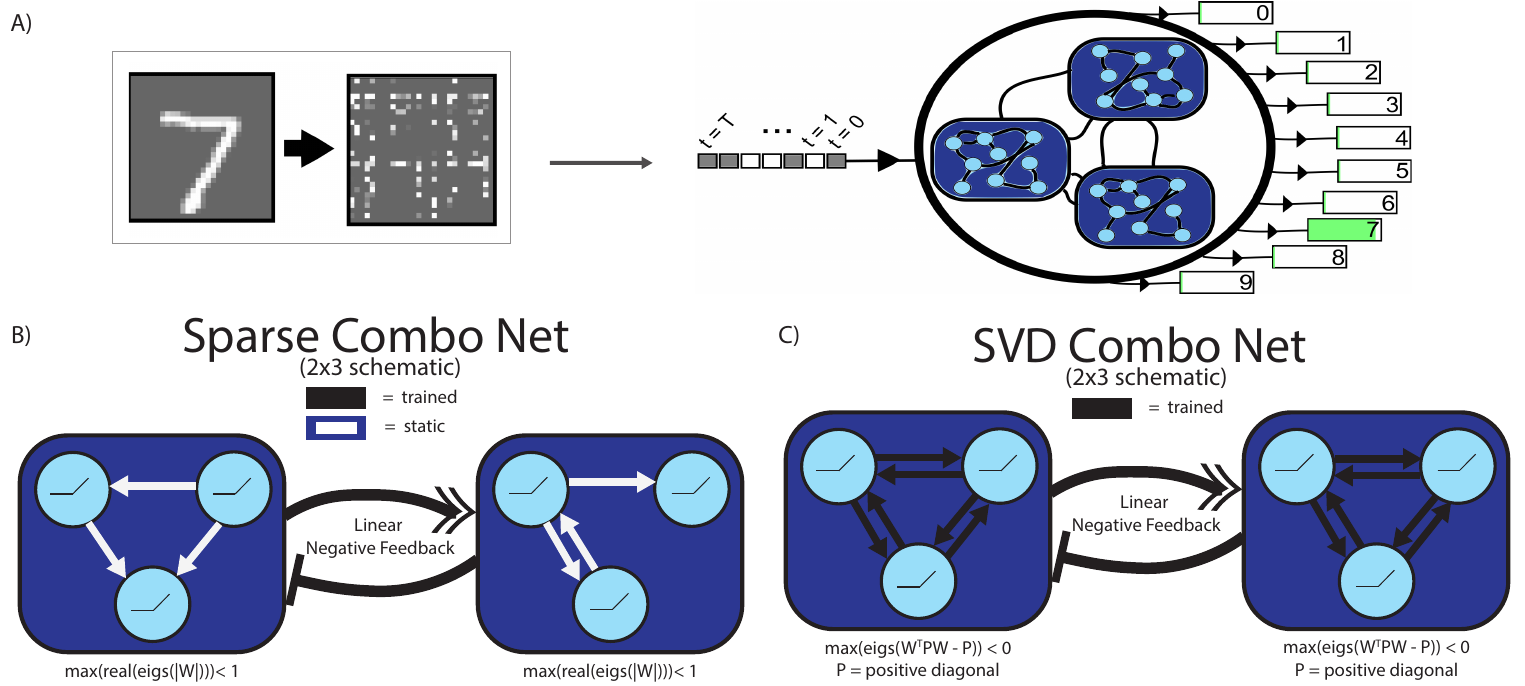}
\caption{\textbf{Summary of task structure and network architectures.} Reproduced from Figure 2 of \citep{kozachkov2022rnns}. In this work, images from CIFAR10 were flattened into an array of pixels and fed sequentially into the modular `network of networks', with classification based on the output at the last time-step. We primarily experimented with modifications to the Sparse Combo Net (B), which was the best performing architecture from \citep{kozachkov2022rnns}, and the one that involved training of only the inter-area weights.}
\label{fig:network-cartoon}
\end{figure}

Extensive experimental details for the pilot Sparse Combo Net were reported in the supplement of \citep{kozachkov2022rnns}, and \citep{ennis2023} goes into even further information about the implementation of this model. However, we provide a brief conceptual summary here to complement our results reported in section \ref{sec:code-results}. An abstract depiction of the stable multi-area RNN network architectures that were tested in practice is reproduced in Figure \ref{fig:network-cartoon}. For the Sparse Combo Net in particular (Figure \ref{fig:network-cartoon}B), white arrows represent the (static) weights that would be found in $\mathbf{W}$ and black arrows represent the (updating) weights that would be found in $\mathbf{L}$.

Note that task input at each timestep was first passed through a (trainable) linear input layer, represented by a weight matrix of size 3 (for RGB CIFAR pixels) by $N$, where $N = p*n$ is the total number of units in the $p \times n$ Combo RNN. On the flip side, the outputs of the Combo RNN neurons at the final timestep in the sequence were used for prediction via passing them through a (trainable) linear output layer. The output layer was represented by a matrix of size $N$ by 10, with 10 being the number of different classes contained in the CIFAR10 dataset. This allowed for computation of cross entropy loss during training based on the true class of each input. At test time, the output index with the maximum value was selected as the predicted label. \\

\noindent To contrast with the experimental results presented in the main text, we also reproduce the summary table of best Sparse Combo Net performance on sequence learning benchmarks, compared against other notable architectures in the literature (Table \ref{table:sota}).  

\begin{table}[h!]
\small
\centering
\begin{tabular}{ | m{2.25cm} || m{0.8cm} | m{0.9cm} || m{1.3cm} | m{1.3cm} | m{1.3cm} || m{0.95cm} | m{0.95cm} | m{0.95cm} | }
\hline
 Name & Stable RNN? & Params & sMNIST \newline Repeats \newline Mean (n) \newline [Min] & psMNIST \newline Repeats \newline Mean (n) \newline [Min] & sCIFAR10 \newline Repeats \newline Mean (n) \newline [Min] & Seq \newline MNIST \newline Best & PerSeq \newline MNIST \newline Best & Seq \newline CIFAR \newline Best \\
\hline\hline
LSTM \citep{chang2019antisymmetricrnn} & & 68K & \centering --- & \centering --- & \centering --- & 97.3\% & 92.7\% & 59.7\% \\
\hline
Transformer \citep{trinh2018cifar} & & 0.5M & \centering --- & \centering --- & \centering --- & 98.9\% & 97.9\% & 62.2\% \\
\hline\hline
Antisymmetric \citep{chang2019antisymmetricrnn} & ? & 36K & \centering --- & \centering --- & \centering --- & 98\% & 95.8\% & 58.7\% \\  
\hline
\rowcolor{Gray}
Sparse Combo Net & \checkmark & 130K & \centering --- & \textbf{96.85\%} (4) \newline [\textbf{96.65\%}] & 64.72\% (10) \newline [63.73\%] & 99.04\% & \textbf{96.94\%} & \textbf{65.72\%} \\  
\hline
Lipschitz \citep{erichson2021lipschitz} & \checkmark & 134K & 99.2\% (10) \newline [99.0\%] & 95.9\% (10) \newline [95.6\%] & \centering --- & \textbf{99.4\%} & 96.3\% & 64.2\% \\  
\hline\hline
CKConv \citep{romero2021ckconv} & & 1M & \centering --- & \centering --- & \centering --- & 99.32\% & 98.54\% & 63.74\% \\
\hline
S4 \citep{gu2022s4} & & 7.9M & \centering --- & \centering --- & \centering --- & \textbf{99.63\%} & \textbf{98.7\%} & \textbf{91.13\%} \\
\hline
Trellis \citep{bai2019trellis} & & 8M & \centering --- & \centering --- & \centering --- & 99.2\% & 98.13\% & 73.42\% \\
\hline
\end{tabular}
\caption{\textbf{Published benchmarks for sequential MNIST, permuted MNIST, and sequential CIFAR10 best test accuracy.} Reproduced from Table 1 of \citep{kozachkov2022rnns} as an external reference point for benchmark task performance. Architectures were grouped into three categories here: baselines, best performing RNNs with claimed stability guarantee, and networks achieving overall SOTA. Within each grouping, networks were ordered by number of trainable parameters (for CIFAR10 if it differed across tasks). The best performing Sparse Combo Net from \citep{kozachkov2022rnns} is highlighted.}
\label{table:sota}
\end{table}

\FloatBarrier

\section{Supplemental Figures}

\begin{figure}[h]
\centering
\includegraphics[width=0.8\textwidth,keepaspectratio]{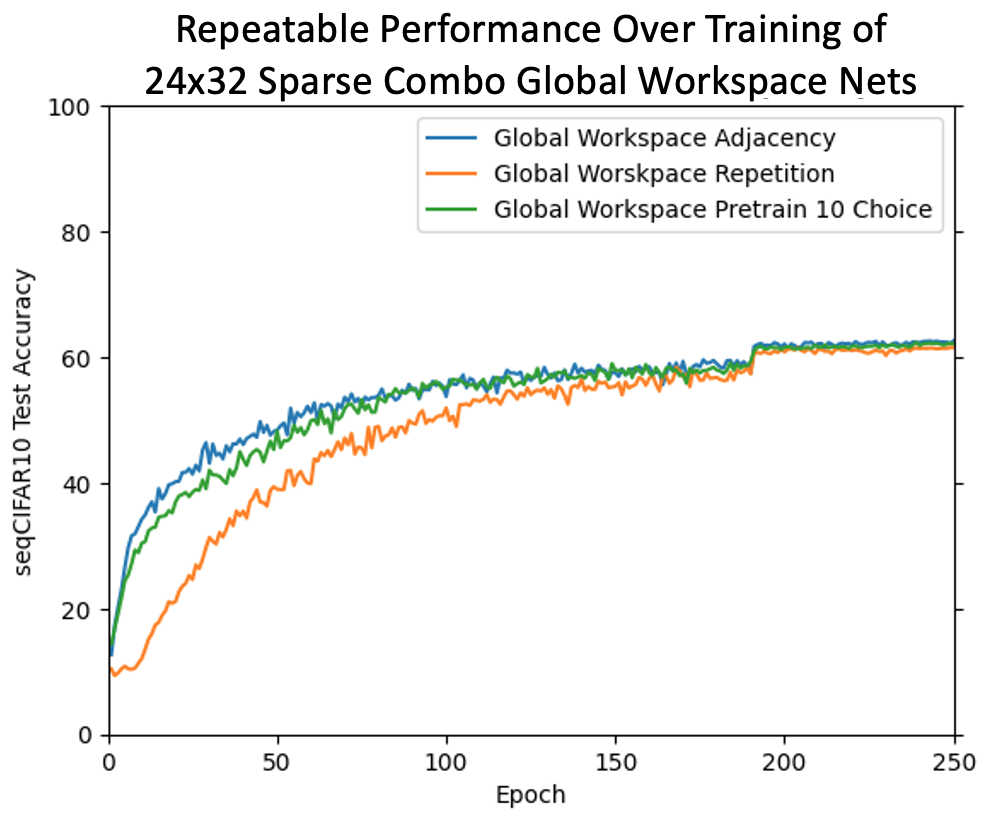}
\caption{\textbf{Global Workspace Sparse Combo Net performance is repeatable across multiple initializations.} Despite the great potential variability found in static nonlinear subnetwork weight initializations, final seqCIFAR10 test accuracy of $24\times32$ Global Workspace Sparse Combo Nets was reliable upon repetition of our methodology. Though there was some uncertainty in early training performance, three different $24\times32$ networks achieved $62.59\%$ (blue), $61.52\%$ (orange), and $62.32\%$ (green) best test accuracies respectively. Their test accuracy over the course of training is plotted here. The final repetition included an initial phase of 10 epochs of all-to-all negative feedback training, which was used to select the subnetwork that would serve as the global workspace in a new network with preserved $\mathbf{W}$, based on mean absolute weight found in $\mathbf{L}$ for that subnetwork at the end of pretraining. This selection mechanism appeared to have no real effect, unsurprising given the limitations demonstrated in Figure \ref{fig:inefficacy-training-all-to-all}. Alternative protocols for selecting a global workspace module most efficiently remain worth exploring however. \newline Note that in the "pretrain select" GW Sparse Combo Net trial depicted here, the selected module was at index 11, as compared to 0 by default. This serves as an unintentional control for any issues with training specific to the lower triangular portion of $\mathbf{B}$ in \eqref{eq:comboRNN}.}
\label{fig:gwt-repeatability}
\end{figure}

\begin{figure}[h]
\centering
\includegraphics[width=\textwidth,keepaspectratio]{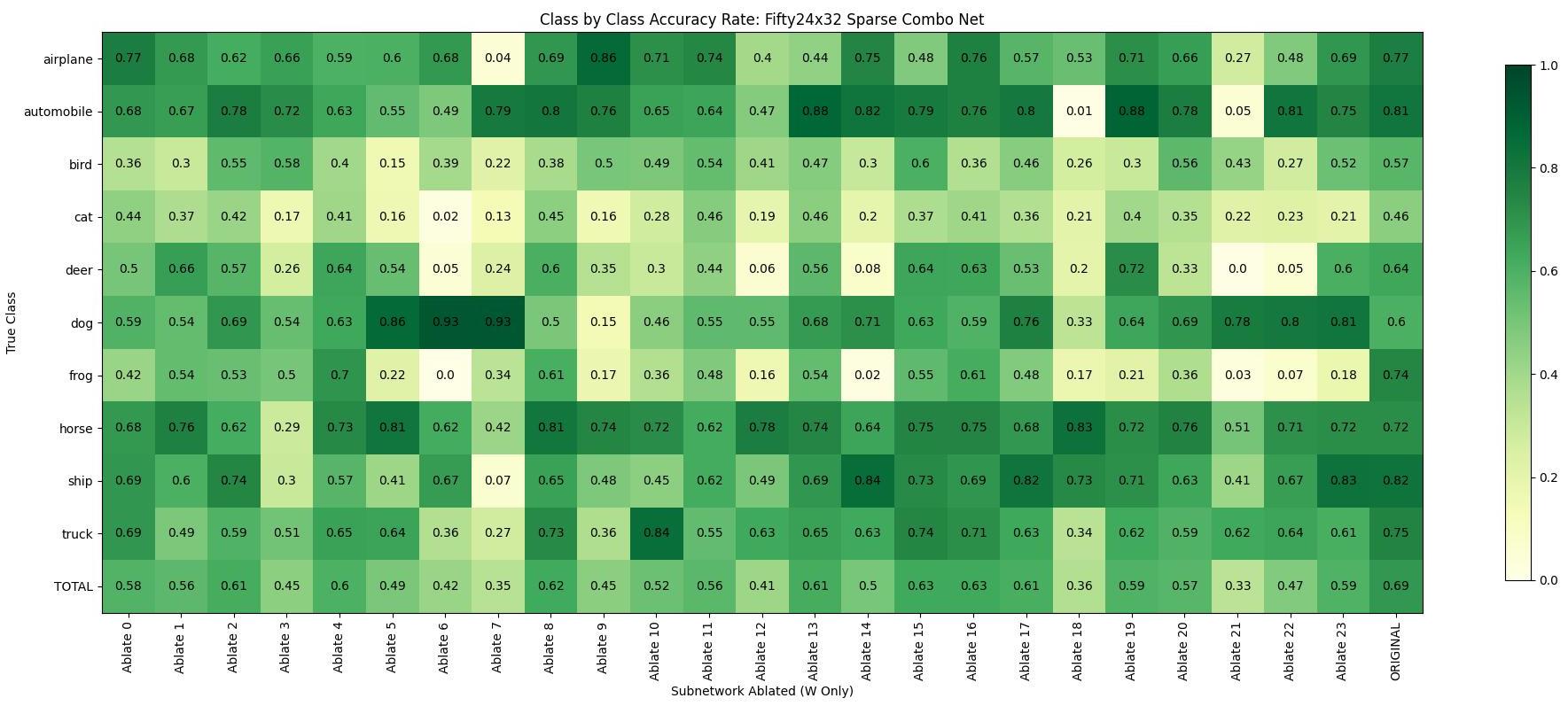}
\caption{\textbf{Class-specific effects of \textit{intra-}subnetwork ablation in the $50\%$ sparse negative feedback $24\times32$ stable multi-area network.} Complements the total subnetwork ablation results depicted for this network in Figure \ref{fig:ablation-fifty}. Here we remove only the weights in the nonlinear internal weight matrix for each subnetwork, represented on the block diagonals of $\mathbf{W}$. \newline This does suggest that benefits of modularity might be somewhat salvaged through modification of subnetwork with minimal changes to the inter-connectivity between it and the larger network, but that does of course limit what modifications would be possible to make.}
\label{fig:ablation-fifty-W}
\end{figure}

\begin{figure}[h]
\centering
\includegraphics[width=\textwidth,keepaspectratio]{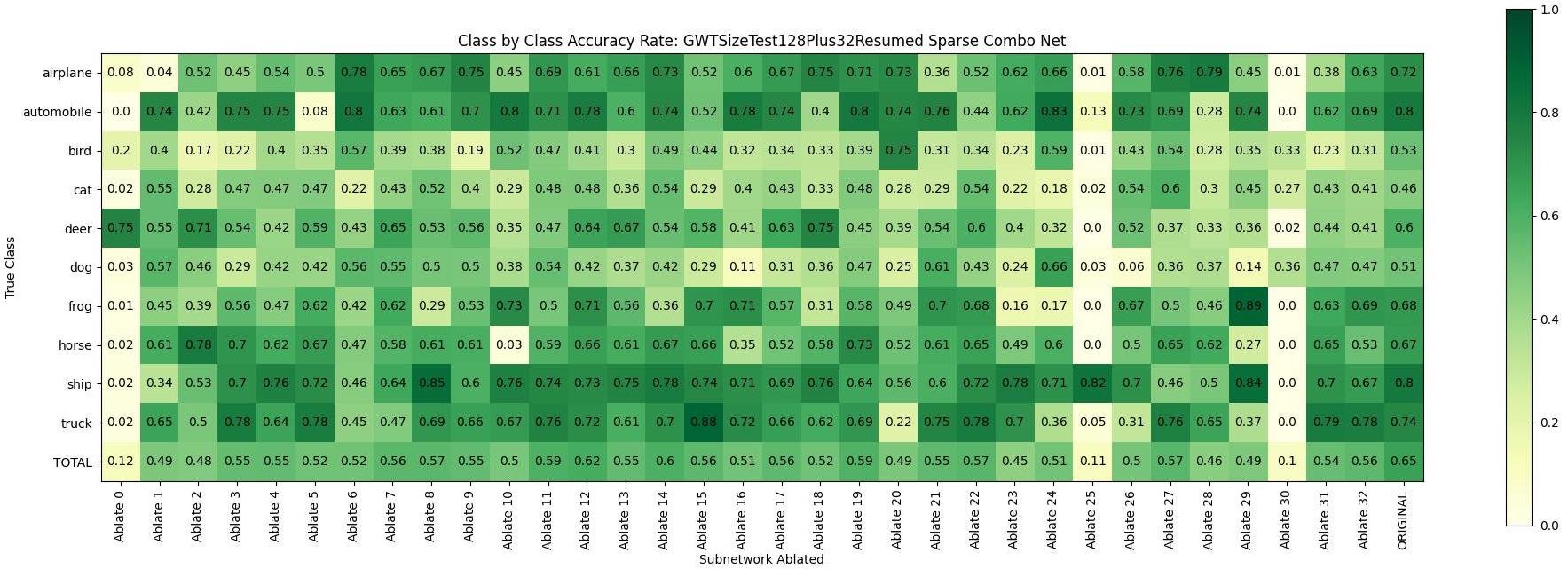}
\caption{\textbf{Class-specific effects of total subnetwork ablation in a larger Global Workspace Sparse Combo Net model.} Using the same figure generation scheme as Figures \ref{fig:class-specific-ablation-impact} and \ref{fig:ablation-fifty}, here we demonstrate the class-specific impacts of total subnetwork removal for the Sparse Combo Net with size 128 Global Workspace and $32\times32$ Contributing Subnet structure (model initially described in Table \ref{table:gwt-comp}). The network was unsurprisingly unable to perform above chance when the GW is ablated, and it was additionally crippled by ablation of subnetworks 25 or 30; but for the most part it was even more robust to subnet removal than the $24\times32$ Global Workspace Sparse Combo Net, with ablation effects even more likely to be focused on prediction of a single class or two. Subnetwork 1 was necessary specifically for prediction of airplanes and to a smaller extent ships. Subnetwork 5 was required for recognition of automobiles. Subnetwork 8 was mostly needed for detection of frogs, while subnetwork 9 was mostly needed for detection of birds. Ablation of subnetwork 10 somewhat impaired a handful of classes, but entirely decimated prediction of horses. Through these observations, we can form hypotheses on what overall network state might look like when images from these different classes are input.}
\label{fig:ablation-gwt-large}
\end{figure}

\FloatBarrier

\section{Extended Experimental Methodology}
\label{sec:methods-supp}
As described in the main text, all experiments were run using the open sourced Sparse Combo Net code from \citep{kozachkov2022rnns} on Google Colab Pro+. Methodology for nonlinear subnetwork initialization and corresponding metric calculation was thus identical to the best performing trials of \citep{kozachkov2022rnns}, as was the base model definition and general training loop. These settings were the same for all trials described in the main text, unless noted otherwise there. Across the whole paper (and all of \citep{kozachkov2022rnns}), Sparse Combo Net was run with $\tau = 1$ and $dt = 0.03$, such that after each timestep (i.e. after each pixel was presented), the current state was updated to be the previous state $+ 0.03 * f(x)$, where $f(x) = -x + \mathbf{W}*\phi(x) + \mathbf{L}*x + u(t)$ with $x$ corresponding to the previous state of the neurons and $u(t)$ corresponding to the values returned by the linear input layer when given the pixel that was just input to the larger network. $\mathbf{W}$ and $\mathbf{L}$ are of course the \emph{intra-} and \emph{inter-} subnetwork weights of \eqref{eq:RNN_of_RNNs} respectively, so they are static throughout this process of evaluating an individual image sequence. Recall that before each new sequence, network starting state was reset.  

We also used the same training function with largely the same hyperparameters as \citet{kozachkov2022rnns}. However due to runtime constraints with larger network sizes, we did increase the training batch size from 64 to 128 for most of our trials, and correspondingly doubled the starting learning rate to 0.002. As the increase in batch size greatly improved runtime, we were able to somewhat extend the training length too, running all of our main trials for 250 epochs instead of the originally intended 200, and accordingly pushing back the two learning rate cut points by 50 epochs each. We used this same training regime for all reported trials. \\

 \noindent Like discussed within Table \ref{table:gwt-comp}, we did use different initialization settings for the global workspace subnetwork when experimenting with differing sizes for this central module. The main difficulty in comparing subnetworks of different sizes for the Sparse Combo Net in general is that sparsity levels cannot be fairly compared -- for a larger nonlinear RNN subnetwork it may be impossible to meet the Theorem 1 condition of \citep{kozachkov2022rnns} without either increasing sparsity or greatly decreasing allowed entry magnitude, but for a smaller nonlinear RNN subnetwork a highly sparse structure might mean that the majority of units have 0 non-zero weights and thus cannot meaningfully contribute. Indeed, there was a point at which increasing subnetwork sparsity decreased the performance of Sparse Combo Nets \citep{kozachkov2022rnns}. As such, we used different sparsity level for initialization of those larger global workspace subnetworks in our GW size experiments. Specifically, we tested a GW with 64 units attached to 24 typically-defined 32 unit subnetworks, and a GW with 128 units attached to 32 typically-defined 32 unit subnetworks. In both cases, the GW subnetwork was initialized with a $1\%$ sparsity level and entry magnitude between $-5$ and $5$. The networks were otherwise trained as before, and their performance on seqCIFAR10 (Table \ref{table:gwt-comp}) was then compared to similarly sized Global Workspace Sparse Combo Nets from the previous trials, which would have GW subnetwork of size 32 with $3.3\%$ sparsity level and entry magnitude between $-6$ and $6$ (as was used for the Sparse Combo Nets more broadly in \citep{kozachkov2022rnns}). 

One additional technical consideration that may warrant further testing in different network topologies is the initialization scheme for inter-area connections. Like \citep{kozachkov2022rnns}, we sample each entry from a normal distribution with mean of 0 and standard deviation $\frac{1}{\sqrt{n}}$ where $n$ is the number of neurons in each subnetwork of a $p \times n$ "RNN of RNNs". In the case where different subnetworks have different sizes, this instead uses the mean number of neurons per subnetwork. With a smaller number of between-subnetwork connections (as in the global workspace topology), one might consider allowing nonzero entries to be initialized to higher entry magnitudes. While in the present paper we have made the most direct comparison possible, this adjustment could allow advantages of the more sparsely connected negative feedback architectures to be even better utilized\footnote{For a more extended set of relevant future experimental directions than provided in this paper, see \citet{ennis2023}.}. \\

\noindent Note that for determining the number of trainable parameters across the different networks, the number of entries updated through training was explicitly verified, in addition to confirming expectations based on model formulas. \citep{kozachkov2022rnns} gives the formula for trainable parameters in an all-to-all negative feedback Sparse Combo Net of size $p \times n$. Here we will give a formula for the Global Workspace Sparse Combo Net trainable parameter count on sequential CIFAR10: $\boxed{14 * p * n + 10 + (p-1) * n^{2}}$

\section{Extended Results}
\label{sec:results-supp}

\subsection{On Pruning for Discovery of Negative Feedback Structures}
\label{subsubsec:pruningA}
\begin{figure}[h]
\centering
\includegraphics[width=\textwidth,keepaspectratio]{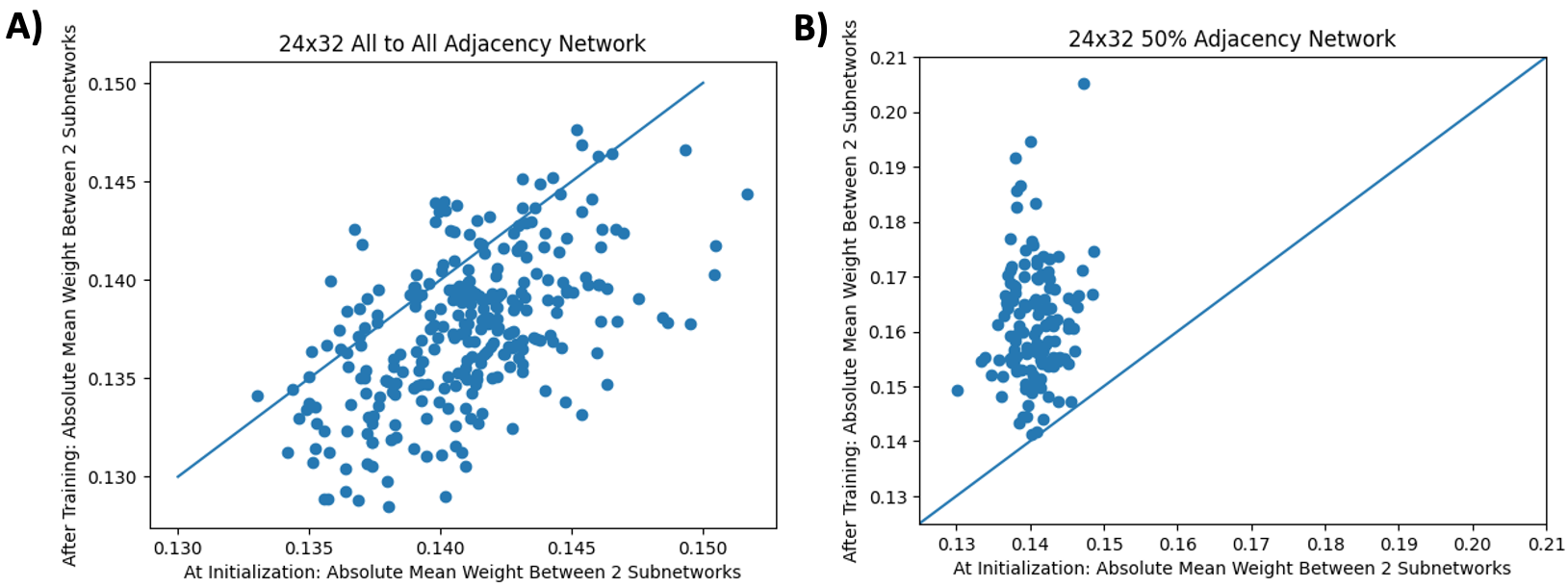}
\caption{\textbf{Final negative feedback connections are much more dependent on initialization when considering all-to-all $\mathbf{A}$ in a larger Sparse Combo Net.} Given the promising results with randomly determined sparsity in the negative feedback adjacency structure, one might hope that early training rounds with $\mathbf{A} = \mathbf{1}$ could identify an even more efficient $\mathbf{A}$ for long term training, via pruning. However, final values of the linear negative feedback connections with full negative feedback, even when trained for an extended period, are not necessarily informative of best network structure. Using $24\times32$ Sparse Combo Nets, we scatter plot absolute mean weight between each pair of subnetworks at initialization versus after training for a $0\%$ sparse $\mathbf{A}$ (A) and for a $50\%$ sparse $\mathbf{A}$ (B), with line $f(x)=y$ for reference. The task performance of the $50\%$ sparse $\mathbf{A}$ network was much better, as can be seen in Table \ref{table:comp}. This also appears to be reflected in how connections change over training, because the strongest subnetwork pairings after training in (B) are significantly stronger than those in (A), as well as much less correlated to strength at initialization.}
\label{fig:inefficacy-training-all-to-all}
\end{figure}

In early trials of pretraining a $24\times32$ all-to-all Sparse Combo Net in order to set a $50\%$ sparse adjacency structure for the same $\mathbf{W}$ on a subsequent training run, results were very similar to the randomly set $50\%$ sparse $\mathbf{A}$ trials. However, this was likely caused by the fact that the $24\times32$ all-to-all negative feedback network struggled to learn in general; even after an entire 250 epochs of training, inter-area weights were largely not higher than at initialization, and their post-training ranking was highly correlated with their ranking at initialization (Figure \ref{fig:inefficacy-training-all-to-all}). For networks with strong performance, including not only the randomly set $50\%$ sparse $\mathbf{A}$ but also the all-to-all $16\times32$, changes in inter-area weights over training appeared significantly more meaningful. 

\noindent Given the results of Figure \ref{fig:inefficacy-training-all-to-all}, it is not a surprise that a standard pruning procedure based on weight magnitude did not confer any benefits over chance when specifying non-zero connections of $\mathbf{A}$. This does suggest alternative methodologies for next experiments though:
\begin{itemize}[itemsep=0mm]
    \item Try ranking subnetwork pairings based on the change of weights in pretraining rather than their magnitude at the end of pretraining.
    \item Try an evolutionary algorithm to test the performance of different adjacency structures after a very short pretraining period. One might consider starting with a global workspace, to be discussed next, as the base structure for building off of.
    \item Perform pruning on $16\times32$ all-to-all Sparse Combo Nets, which learn much better, and then combine the results to create a larger e.g. $32\times32$ Sparse Combo Net for training.
    \item Perform the pretraining on a $50\%$ sparse $\mathbf{A}$ $24\times32$ network instead, and then try to strategically prune the adjacency structure to e.g. $25\%$ non-zero connections based on that. 
\end{itemize}

\FloatBarrier

\subsection{Characterizing Network Test Performance}

\begin{figure}[h]
\centering
\includegraphics[width=\textwidth,keepaspectratio]{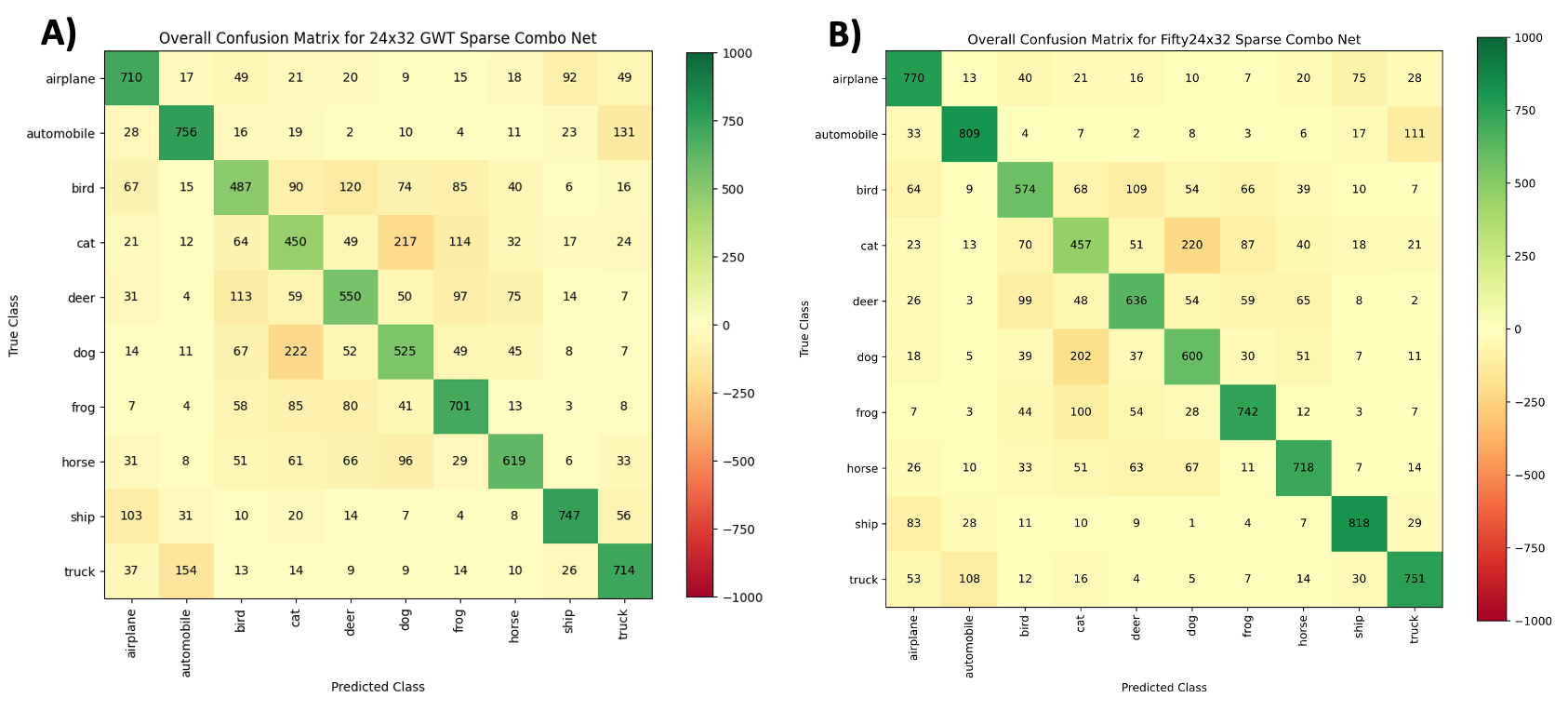}
\caption{\textbf{Class-specific accuracy and common mistakes share a similar structure across different multi-area RNN topologies.} Confusion matrices for final seqCIFAR10 test performance from the $24\times32$ Global Workspace (A) and Random $50\%$ of Negative Feedback Pairings (B) Sparse Combo Nets listed in Table \ref{table:comp}. Diagonal entries represent correct predictions for each possible class and are colored from pale yellow representing the worst count of 0 to dark green representing the best count of 1000. Off-diagonal entries $i,j$ represent the number of images of class $i$ that were incorrectly labeled as class $j$, and are colored from pale yellow representing the best count of 0 such mix-ups, to dark red representing that all 1000 $i$ images in the test set were incorrectly labeled as $j$. The 10 classes of CIFAR10 are depicted here in alphabetical order. The performance of the $50\%$ sparse $\mathbf{A}$ network (B) was overall superior, but generally came through small improvements across all weaknesses rather than any major perspective shift. Cat, bird, dog, and deer (in order) were the worst performing classes in both confusion matrices, the only classes below $60\%$ in (A) and the only classes below $70\%$ in (B). Additionally, all vehicles were easier than all animals, and dog/cat mix-ups were bidirectionally the largest type of mistake by far. A similar trend was repeated across the confusion matrices for all networks reported in Table \ref{table:comp}, as well as the Global Workspace repetitions of Figure \ref{fig:gwt-repeatability}.}
\label{fig:main-confusion-matrices}
\end{figure}

To identify possible subtasks of relevance for the seqCIFAR10 benchmark, it will be important to determine what these networks currently perform well versus poorly at, and whether that may differ by network structure. It turned out that confusion matrices demonstrated similar trends across our trials (Figure \ref{fig:main-confusion-matrices}), so there is not an obvious combination of existing networks to perform. However, these results indeed highlight plausible next steps for directed subtask training.

One unsurprising commonality across confusion matrices (Figure \ref{fig:main-confusion-matrices}) was that misclassification of vehicles largely involved other vehicles and misclassification of animals largely involved other animals. Amongst vehicles, distinction of automobiles and trucks was most problematic, though animal classification consistently presented the greatest challenges. Confusion of dogs and cats, and perhaps less expectedly deer and birds, were the largest sources of mistakes. A straightforward next step then would be to independently train one Sparse Combo Net on animal classification only and another on vehicle classification only, and then train their combination on the complete seqCIFAR10 task. To take an even more hierarchical training approach, the animal network might include a set of subnetworks already pretrained on just distinguishing cats and dogs or just distinguishing deer and birds. 

\FloatBarrier

\subsection{Relationships Between Fixed Adjacency Structure and Learned Weights}

\subsubsection{Randomized Sparse Adjacency}
\begin{figure}[h]
\centering
\includegraphics[width=\textwidth,keepaspectratio]{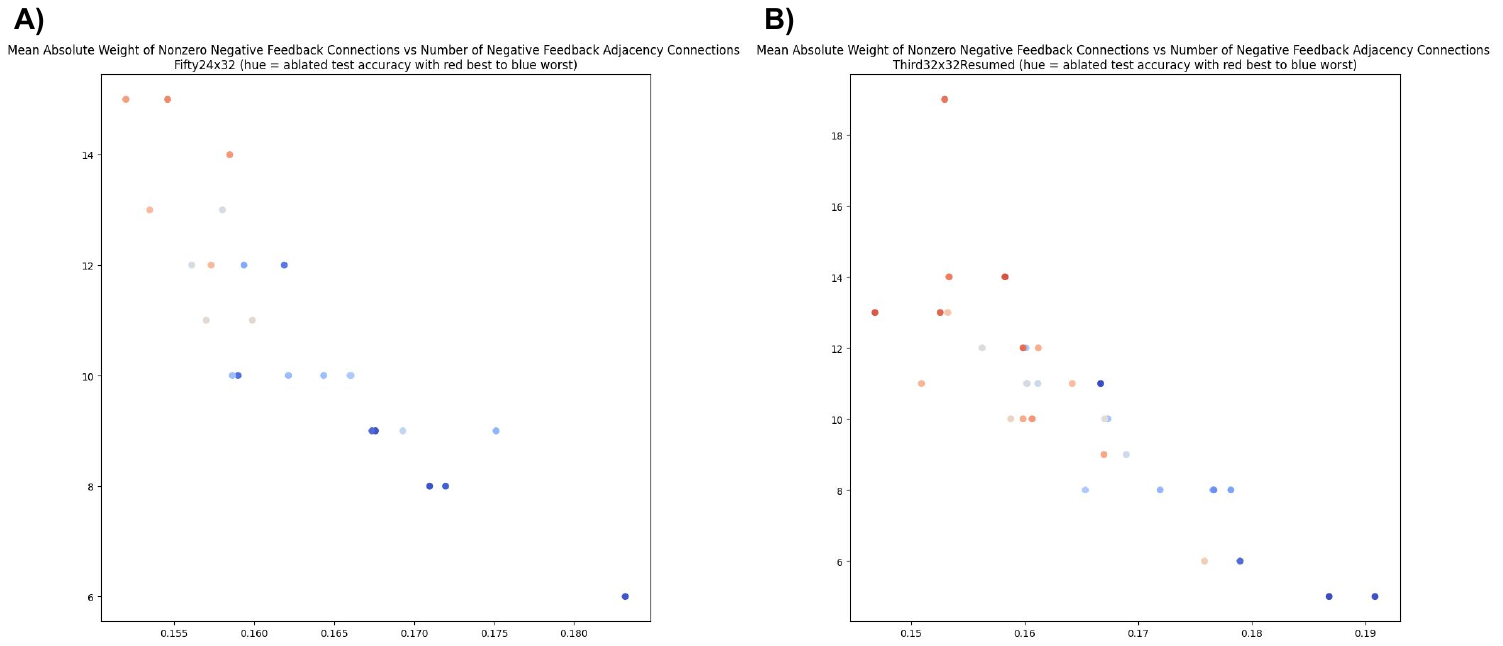}
\caption{\textbf{Subnetworks randomly initialized with fewer inter-areal connections learn higher magnitude weights and are more critical to network performance.} In the sparsified negative feedback adjacency experiments, which pairs of subnetworks would be connected was determined randomly at initialization. For both the $50\%$ non-zero $\mathbf{A} 24\times32$ Sparse Combo Net (A) and the $33.3\%$ non-zero $\mathbf{A} 32\times32$ Sparse Combo Net (B), ablating subnetworks that happened to be joined in negative feedback with a smaller number of other subnetworks was consistently more devastating to test performance on seqCIFAR10. Moreover, those subnetworks with lesser inter-area connectivity tended to learn higher magnitude negative feedback weights for their non-zero entries within $\mathbf{L}$. Here we scatter plot the subnetworks of both networks, plotting the mean absolute magnitude of their linear \emph{inter}-area weights (x) against the number of other subnetworks they were connected to via $\mathbf{A}$ (y). The hue of each point is the test accuracy of the network with that subnetwork ablated, from dark blue for the worst performance to dark red for the best ('coolwarm' colormap).}
\label{fig:rand-adjacency-connectivity}
\end{figure}

\begin{figure}[h]
\centering
\includegraphics[width=\textwidth,keepaspectratio]{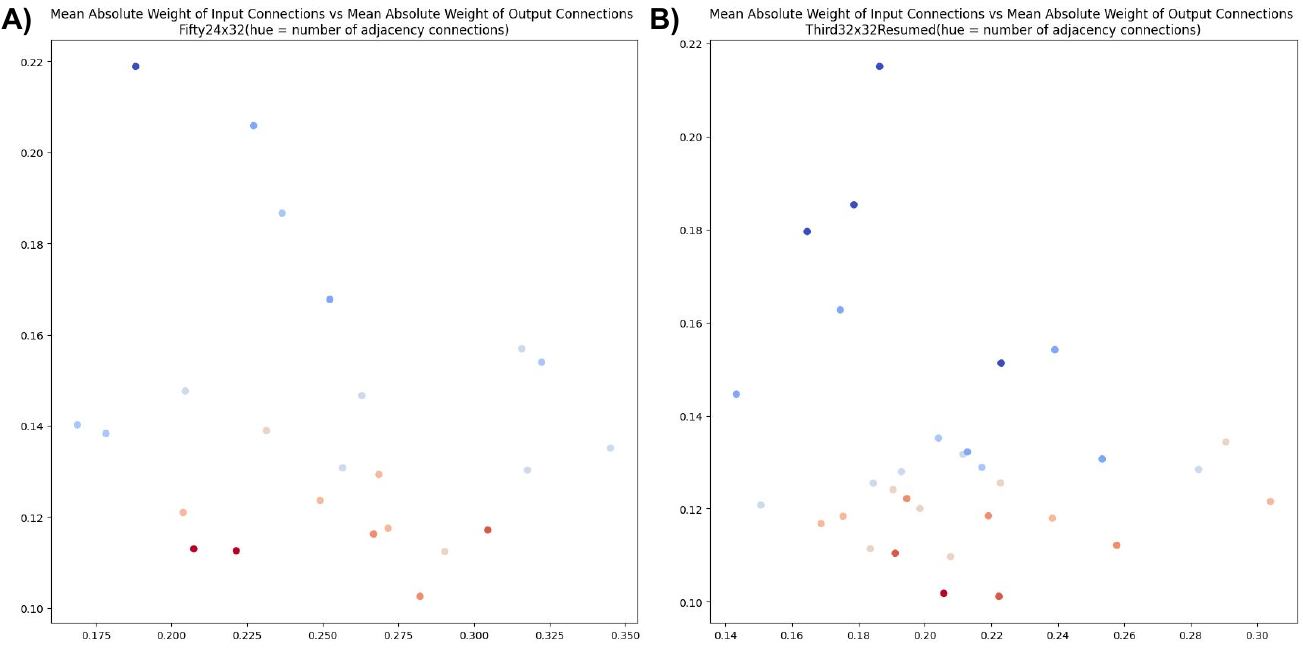}
\caption{\textbf{Subnetworks randomly initialized with fewer inter-areal connections learn higher magnitude final output weights but possibly lower magnitude image input weights.} Here we scatter the mean absolute weight of subnetwork-associated input connections (x-axis) against the mean absolute weight of subnetwork-associated output connections (y-axis) for both the $50\%$ negative feedback $24\times32$ Sparse Combo Net (A) and the $33.3\%$ negative feedback $32\times32$ Sparse Combo Net (B). In both panels points are colored according to the number of nonzero negative feedback connections in the adjacency for that subnetwork, from lowest (blue) to highest (red). In both cases, higher output weights were consistently associated with those subnetworks that were randomly initialized to have fewer total \emph{inter-}area connections. At the most extreme output magnitudes, we also observed lower input weight magnitudes.}
\label{fig:rand-adjacency-io}
\end{figure}

To elucidate properties of learned network structure in the random sparsified negative feedback adjacency experiments, we also looked for relationships between static and learned properties of subnetwork weights as well as interactions between these relationships and the importance of the subnetwork for overall performance. A recurring theme with the random adjacency was that subnetworks initialized to have fewer negative feedback pairings with other subnetworks ended up with stronger non-zero inter-area weights and were actually the most impactful subnetworks to ablate at test time (Figure \ref{fig:rand-adjacency-connectivity}). Similarly, we observed a relationship between adjacency connectivity and the strength of learned input and especially output weights (Figure \ref{fig:rand-adjacency-io}).

An obvious change to help these networks learn more efficiently would thus be to change the range of possible $\mathbf{L}$ initialization values based on the subnetwork associated with the entry, allowing less connected subnetworks to start with higher magnitude inter-area feedback weight values. This also raises the question of whether the observed pattern is conducive to learning in general or only an adaptation to handle the unevenness of the random adjacency scheme. Comparison with an evenly distributed $50\%$ sparse adjacency is therefore another easy follow-up experiment. 

\FloatBarrier

\subsubsection{Global Workspace Adjacency}
\begin{SCfigure}[1][h]
\centering
\includegraphics[width=0.5\textwidth,keepaspectratio]{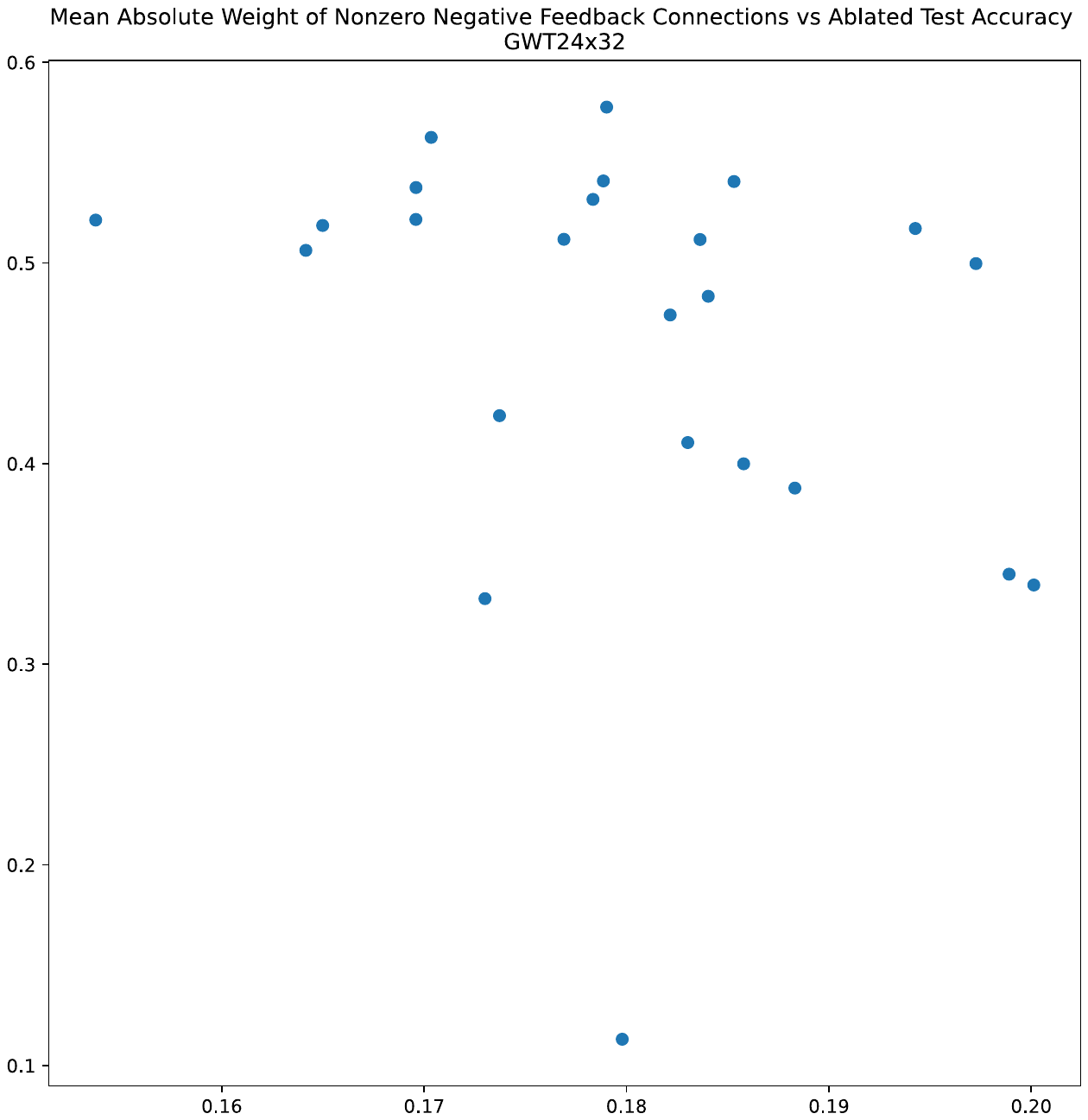}
\caption{\textbf{Relationship between connection strength to the global workspace and negative impact of post-training subnetwork removal.} Here we scatter plot the subnetworks of the $24\times32$ Global Workspace Sparse Combo Net, plotting the mean absolute magnitude of their linear \emph{inter}-area weights (x) against the seqCIFAR10 test accuracy of the network with that subnetwork ablated (y). Note that for 23 of the 24 depicted points, their negative feedback weights consist solely of weights to/from the global workspace. Of course the GW weights include connections to/from all other subnetworks, so this module will naturally fall in the middle of mean negative feedback weight magnitude, as well as have extremely low ablated test accuracy.}
\label{fig:gwt-ablation-connectivity}
\end{SCfigure}

To elucidate properties of learned network structure in the global workspace (GW) negative feedback adjacency experiments, we also looked for relationships between static and learned properties of subnetwork weights as well as interactions between these relationships and the importance of the subnetwork for overall performance. In the case of the GW models, the average magnitude of inter-area weights (in this case just to/from the global workspace module) was again somewhat predictive of subnetwork ablation impact (Figure \ref{fig:gwt-ablation-connectivity}). As only a handful of subnetworks were highly important to performance in GW networks, this could serve as a useful heuristic for identifying those subnetworks that are most safe for downstream modification. Conversely, it may help us identify subnetworks for further study, to better understand what information those modules are contributing to the GW.  

\begin{figure}[h]
\centering
\includegraphics[width=\textwidth,keepaspectratio]{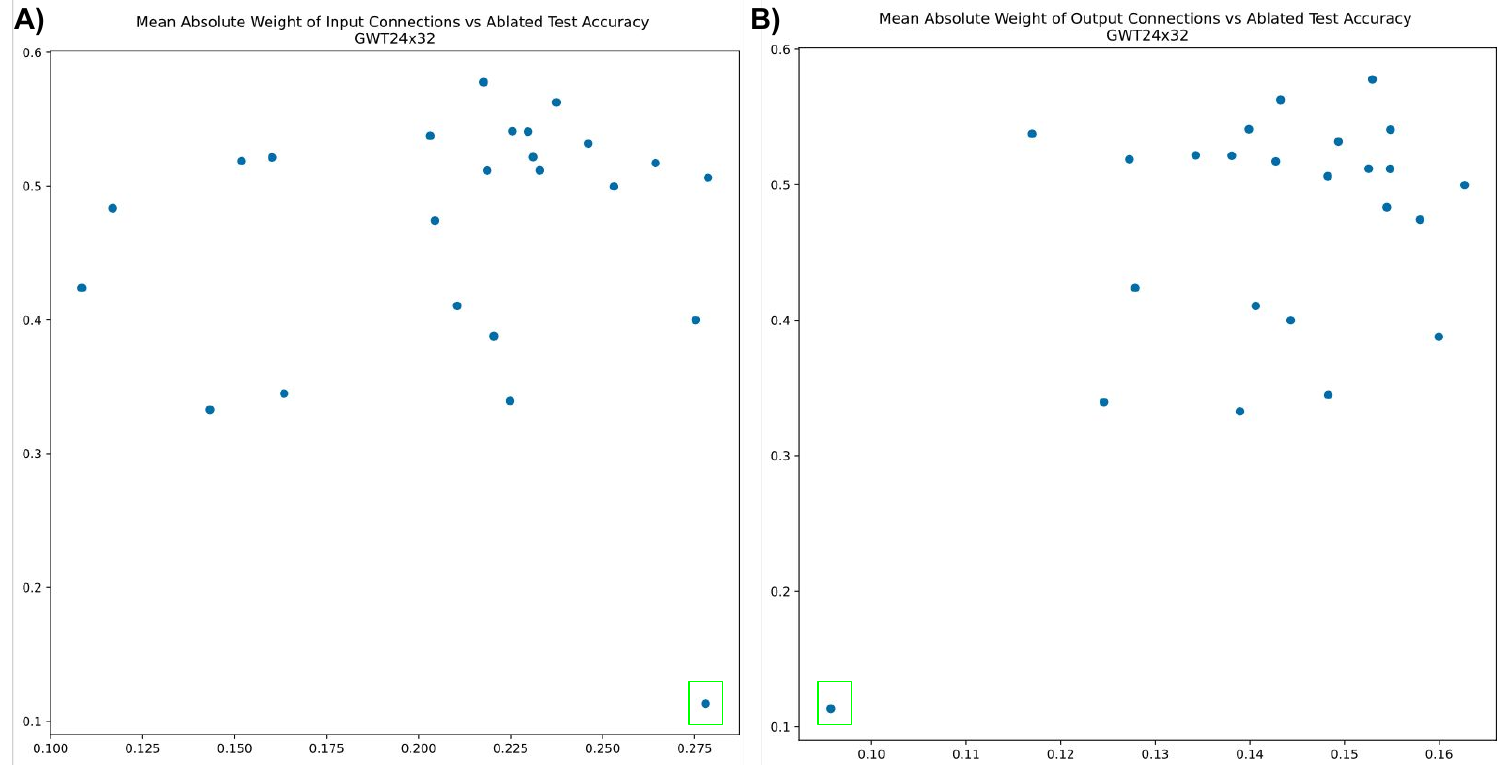}
\caption{\textbf{Trained Global Workspace Sparse Combo Nets tend to have GW module with high magnitude input weights and low magnitude output weights.} Here we show for the $24\times32$ Global Workspace Sparse Combo Net the mean absolute weight of subnetwork-associated input connections (x-axis) versus the test accuracy of the network (y-axis) with that subnetwork totally ablated (A), and similarly the mean absolute weight of subnetwork-associated output connections versus the test accuracy of the network with that subnetwork totally ablated (B). In both panels the point representing the GW subnetwork is boxed.}
\label{fig:gw-io}
\end{figure}

Another consistent finding in the Global Workspace Sparse Combo Nets was a very high average input weight magnitude and a very low average output weight magnitude specifically for GW units (Figure \ref{fig:gw-io}). This suggests that the Global Workspace performs its own computations on the input pixels in addition to receiving signals from other component subnetworks, and both are highly important to the signals that it communicates back to the other subnetworks. However despite being a centralized node, the GW appears to have a much more minor role in communicating the final output, as the states of most other subnetworks are more strongly weighted.

\FloatBarrier

\subsection{Additional Ablation Tests}

\begin{figure}[h]
\centering
\includegraphics[width=\textwidth,keepaspectratio]{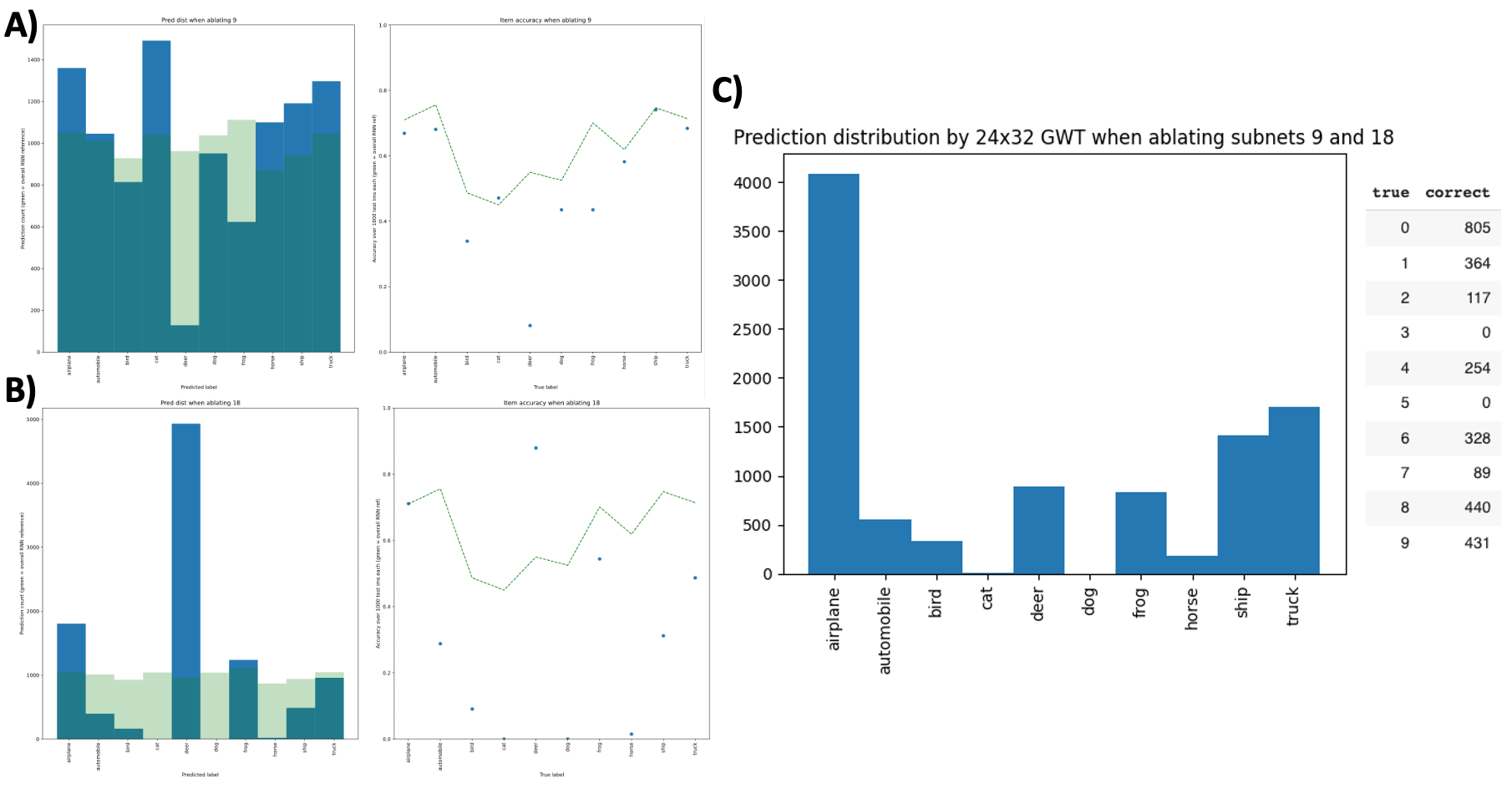}
\caption{\textbf{Dual ablation of subnetworks with potentially competing roles in the Global Workspace model.} When ablating subnetwork 9 of the $24\times32$ Global Workspace Sparse Combo Net, the primary effect was loss of prediction of the class "deer" (A, left), with accuracy rate on images of deer obviously taking the largest hit compared to the unperturbed network (A, right). When ablating subnetwork 18, the network never predicted cat or dog, and also had large accuracy drops on horse, bird, automobile, and ship (B). For most images of animals, it guessed "deer". Given the nearly opposite profiles seen here, one might wonder the effects of ablating both these subnetworks simultaneously. The distribution of guesses when both 9 and 18 were removed is plotted here (C), along with an inset table of the number of correct predictions for each true class label in the seqCIFAR10 testing dataset (compare against 1000 of each total). While the combined ablation brings the rate of prediction of "deer" to a normal level, the accuracy of those predictions is quite low with only 254 true positives. At the same time, the classes ignored when ablating 18 alone remain less frequently guessed, and neither cat nor dog are ever predicted in the dual ablated case. Instead, the distribution is shifted more towards "airplane", indicating that weaknesses when ablating subnetworks 9 and 18 individually were largely additive -- thus leaving the dual ablation network with not only poor results across all animal labels, but also poor performance in even distinguishing animals from vehicles.
\newline From output ablation results (Figure \ref{fig:class-specific-ablation-impact}D), we can also see that predictions of deer were largely reliant on the final state of subnetwork 9, while predictions of dog or cat (and to a lesser extent horse) were reliant on the final state of subnetwork 18. Interestingly, removing image input to subnets 9 or 18 did not produce similar results (Figure \ref{fig:class-specific-ablation-impact}C), but targeted ablation of internal weights $\mathbf{W}$ did (Figure \ref{fig:class-specific-ablation-impact}B), indicating that the readouts of 9/18 were reliant on internal computation of signal from the global workspace. This helps to contextualize the results of dual total ablation of these two subnetworks found in this figure, as from the systematic ablation view it is clear that removal of 18 results in over-prediction of deer due to the inability to predict cat or dog without it, rather than due to some sort of "not deer" signal.}
\label{fig:testing-targeted-dual-ablation}
\end{figure}

\FloatBarrier

\subsection{The Global Workspace with Internal Negative Feedback}
In our early experiments, increased global workspace size improved performance (Table \ref{table:gwt-comp}). Another alternative for the GW module that might further improve sequence learning would be to use a small Sparse Combo Net consisting of all-to-all negative feedback between a handful of smaller nonlinear RNN subnetworks, instead of using a single larger nonlinear RNN subnetwork. Here a $36\times32$ Sparse Combo Net with the first 4 subnetworks connected in negative feedback to all other subnetworks could be compared directly against the Size 128 Global Workspace with $32\times32$ Subnets (Table \ref{table:gwt-comp2}). 

\begin{table}[h!]
\centering
\begin{tabular}{ | m{8.25cm} || m{1.6cm} | m{1.9cm} | m{1.2cm} | }
\hline
 Name & Trainable Params Count & seqCIFAR10 \newline Test \newline Accuracy & Final \newline Train \newline Loss \\
\hline\hline
26x32 Global Workspace Sparse Combo Net \newline (i.e. Size 32 Global Workspace With 25x32 Subnets) & 37,258 & 61.07\% & 0.93 \\
\hline
Size 64 Global Workspace With 24x32 Subnets & 60,810 & 63.72\% & 0.6806 \\
\hline
\rowcolor{Gray}
2x32 Global Workspace With 24x32 Subnets & 61,834 & 63.31\% & 0.7529 \\
\hline\hline
Size 128 Global Workspace With 32x32 Subnets & 147,210 & 65.57\% & 0.5532 \\
\hline
\rowcolor{Gray}
4x32 Global Workspace With 32x32 Subnets & 153,354 & 64.75\% & 0.6546 \\
\hline
\end{tabular}
\caption{\textbf{Performance with larger GW modules decreased when implemented as a mini Sparse Combo Net rather than as a single large nonlinear RNN.} The same stats as in Table \ref{table:gwt-comp} are presented here, now comparing those trials of a larger global workspace subnetwork with parallel trials of a global workspace made up from multiple smaller individual subnetworks (highlighted rows). Additionally, to make a one-to-one comparison in terms of number of units with a single small GW subnetwork, a trial of a $26\times32$ Global Workspace Sparse Combo Net as defined before was run. Implementing a larger GW module as a mini Sparse Combo Net did not confer any performance boost over the more straightforward implementation of a larger GW, though it did appear to improve accuracy above what could be achieved by a small GW.}
\label{table:gwt-comp2}
\end{table}

\begin{SCfigure}[1][h]
\centering
\includegraphics[width=0.5\textwidth,keepaspectratio]{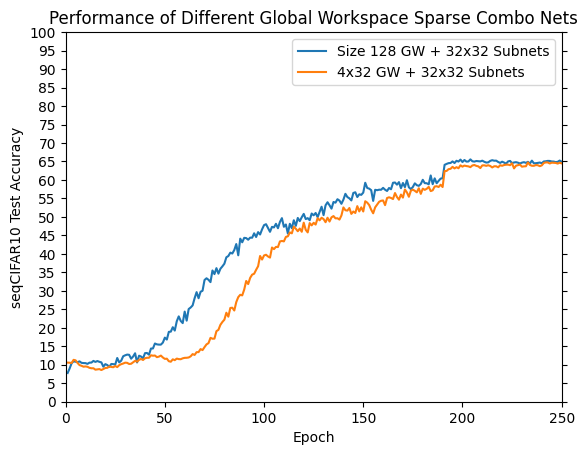}
\caption{\textbf{Sequential CIFAR10 test accuracy over training for two different global workspace structures in a GW Sparse Combo Net.} Using 32 contributing nonlinear subnetworks of 32 units each, independently randomly initialized using the protocol described in \citep{kozachkov2022rnns}, we test different possible global workspace structures to train inter-area connections. One was a size 128 fixed nonlinear GW (described in Table \ref{table:gwt-comp}), which then had negative feedback connections trained between it and all contributing subnetworks (blue). The other was a $4 \times 32$ all-to-all negative feedback Sparse Combo Net, which then also had negative feedback connections trained between it an all contributing subnetworks (orange). Here we plot the test accuracy over training for both cases; final test accuracies were similar, but including negative feedback to train within the GW slowed early training.}
\label{fig:size36progressions}
\end{SCfigure}

\begin{SCfigure}[1][h]
\centering
\includegraphics[width=0.5\textwidth,keepaspectratio]{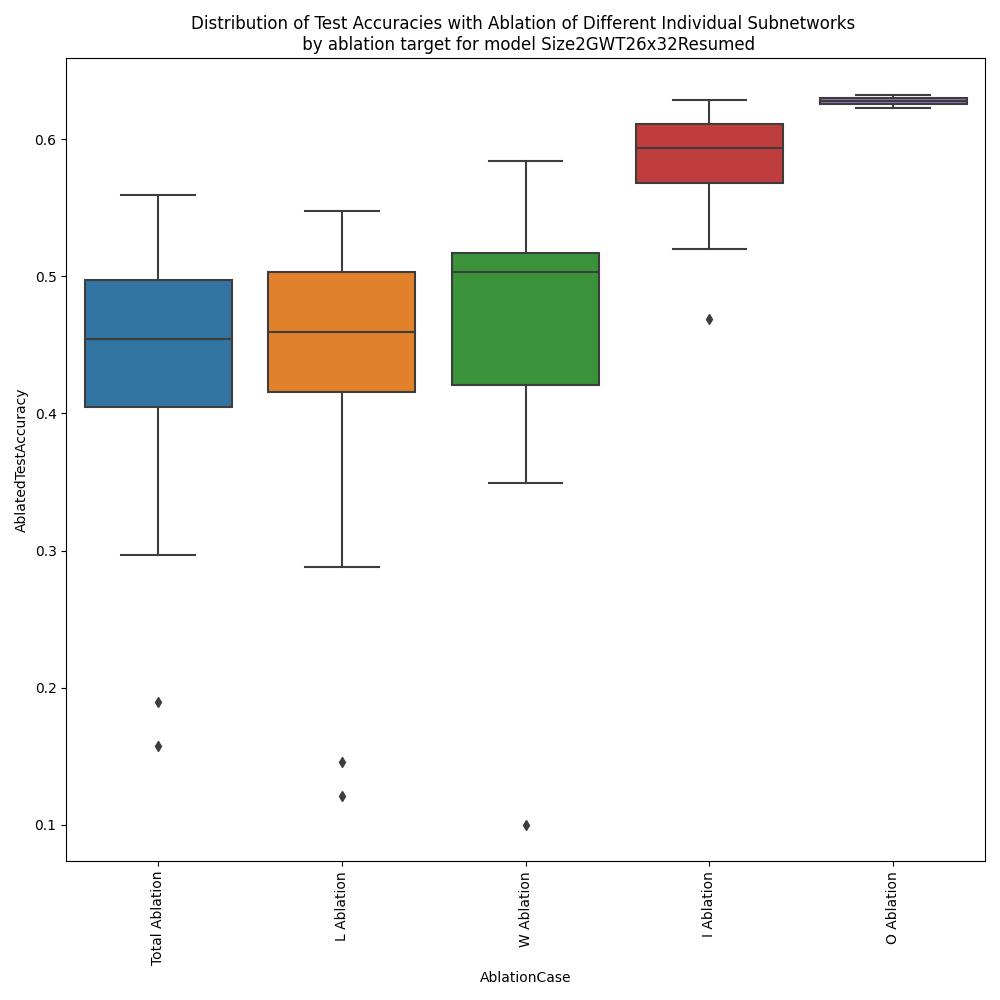}
\caption{\textbf{Subnetworks randomly initialized with fewer inter-areal connections are more critical to network performance in the $50\%$ sparse negative feedback $24\times32$ stable multi-area network.} Using the same figure generation scheme as Figure \ref{fig:ablation-by-type}, we now show the effects of different types of ablation on the subnetworks in a GW Sparse Combo Net with linear negative feedback connections found inside the GW (i.e. a $2 \times 32$ Sparse Combo Net serving as GW for $24 \times 32$ contributing subnetworks). Impact by type of ablation showed a similar trend in this case, but performance was overall moderately more susceptible to ablation than in the case with a typical global workspace.}
\label{fig:Size2GWT26x32Resumed}
\end{SCfigure}

However, the networks with GW implemented as a mini Sparse Combo Net did not outperform those with the GW implemented as a single fixed nonlinear RNN (Table \ref{table:gwt-comp2}), and showed potential limitations related to training speed (Figure \ref{fig:size36progressions}) and robustness to subnetwork ablation (Figure \ref{fig:Size2GWT26x32Resumed}). It is unclear to what extent this experiment speaks directly to graph structure suitability or perhaps the role of nonlinear connections in the GW, versus to the importance of weights within the GW remaining static during final task training. These questions would be straightforward to address with future experimentation though -- particularly through the use of our novel theoretical results on GW multi-area network stability conditions (section \ref{subsec:gw-math}), which allow for nonlinear connections between subnetworks unlike the more general "RNN of RNNs" case presented in \cite{kozachkov2022rnns}.

\FloatBarrier

\end{document}